\documentclass[lettersize,journal]{IEEEtran}
\usepackage{amsmath,amsfonts}
\usepackage{algorithm}
\usepackage{algorithmicx}
\usepackage{algpseudocode}
\usepackage{array}
\usepackage[caption=false,font=normalsize,labelfont=sf,textfont=sf]{subfig}
\usepackage{textcomp}
\usepackage{stfloats}
\usepackage{url}
\usepackage{verbatim}
\usepackage{graphicx}
\usepackage{cite}
\usepackage{multirow}
\usepackage{booktabs}
\usepackage{color}
\usepackage{siunitx}
\usepackage{amssymb}
\begin{document}

\title{Context Recovery and Knowledge Retrieval: \\A Novel Two-Stream Framework \\for Video Anomaly Detection}

\author{Congqi Cao\textsuperscript{* \textdagger}, Yue Lu\textsuperscript{*}, and Yanning Zhang
	\thanks{\textsuperscript{*}Equal contribution. \textsuperscript{\textdagger}Corresponding author.}
	\thanks{Congqi Cao, Yue Lu and Yanning Zhang are with the National Engineering Laboratory for Integrated Aero-Space-Ground-Ocean Big Data Application Technology, School of Computer Science, Northwestern Polytechnical University, Xi'an 710129, China (e-mail: congqi.cao@nwpu.edu.cn; zugexiaodui@mail.nwpu.edu.cn; ynzhang@nwpu.edu.cn).}
}


\maketitle

\begin{abstract}

Video anomaly detection aims to find the events in a video that do not conform to the expected behavior.
The prevalent methods mainly detect anomalies by snippet reconstruction or future frame prediction error.
However, the error is highly dependent on the local context of the current snippet and lacks the understanding of normality.
To address this issue, we propose to detect anomalous events not only by the local context, but also according to the consistency between the testing event and the knowledge about normality from the training data.
Concretely, we propose a novel two-stream framework based on context recovery and knowledge retrieval, where the two streams can complement each other.
For the context recovery stream, we propose a spatiotemporal U-Net which can fully utilize the motion information to predict the future frame.
Furthermore, we propose a maximum local error mechanism to alleviate the problem of large recovery errors caused by complex foreground objects.
For the knowledge retrieval stream, we propose an improved learnable locality-sensitive hashing, which optimizes hash functions via a Siamese network and a mutual difference loss.
The knowledge about normality is encoded and stored in hash tables, and the distance between the testing event and the knowledge representation is used to reveal the probability of anomaly.
Finally, we fuse the anomaly scores from the two streams to detect anomalies.
Extensive experiments demonstrate the effectiveness and complementarity of the two streams, whereby the proposed two-stream framework achieves state-of-the-art performance on four datasets.

\end{abstract}


\section{Introduction}
\IEEEPARstart{V}{ideo} anomaly detection (VAD) is the task of detecting the events in a video that do not conform to the expected behavior \cite{SurveySingleScene2022ramachandra}, which has wide applications in intelligent surveillance and public security.
It is an extremely challenging task for the following reasons.
First, anomalous events rarely occur and their categories are agnostic and unbounded.
In most practical application scenarios, we can only obtain the normal data, while the abnormal data is absent.
Second, video anomalies are scene-dependent \cite{SurveySingleScene2022ramachandra}.
For example, playing football is normal on the pitch but anomalous on the road.
Third, some kinds of normal events happen frequently while some happen occasionally.
If an algorithm cannot handle the imbalanced data distribution well, it is easy to treat infrequent normal events as abnormal events, resulting in false positives.

In the early works, researchers use one-class support vector machine (OC-SVM) \cite{LearningDeep2015xua, DeepAppearance2017smeureanu, ObjectCentricAutoEncoders2019ionescua}, KNN \cite{VideoAnomaly2012saligramaa, StreetScene2020ramachandraa}, clustering \cite{DetectingAbnormal2019ionescua, ClusterAttention2020wanga, GraphEmbedded2020markovitza, ClusteringDriven2020chang}, a mixture of Gaussians \cite{GaussianProcess2015cheng, VideoAnomaly2020fana} and other methods to represent normal events with representative features, such as clustering centers and distribution parameters.
The anomaly probability of a testing event is determined by the distance between it and the representative features.
Although these methods have the advantage of good interpretability, they lack adequate and flexible generalization of normal events.
For example, the number of cluster centers needs to be set manually and is usually small.
With the development of deep learning, methods based on snippet reconstruction and future frame prediction \cite{FutureFrame2018liua, MemorizingNormality2019gonga, ClozeTest2020yu, BMANBidirectional2020lee, LearningMemoryGuided2020parka, FutureFrame2021luo, SmithNetStrictness2021nguyen, LearningNormal2021lva, HybridVideo2021liua, AppearanceMotionMemory2021caia, RobustUnsupervised2021wanga, MultiEncoderEffective2021fang, AnomalyDetection2022fang, VariationalAbnormal2022li, SelfSupervisedAttentive2022huang, InfluenceawareAttention2022zhang} have become popular in recent years.
These two kinds of methods train auto-encoders to reconstruct the current snippet or predict the future frame, and calculate the anomaly probability by the reconstruction or prediction error.
Since they both aim to recover the context information of frames, we classify them as the context recovery method.
This method is good at distinguishing short-term anomalous movements, but lacks the understanding of normality.
For example, a snippet reconstruction model can accurately reconstruct the action of playing football not only on the pitch but also on the road, making it impossible to detect the anomalous event.
Although some memory-augmented context recovery methods \cite{MemorizingNormality2019gonga, LearningMemoryGuided2020parka, LearningNormal2021lva, HybridVideo2021liua, AppearanceMotionMemory2021caia} try to explicitly utilize the diversity of normal data, they are essentially designed for recovering the context.
Therefore, it is still difficult for such models to mitigate that drawback.
In addition, the data-driven nature of deep neural networks makes it hard to reconstruct the normal events that seldom occur.
To solve the above problems, it is necessary to make full use of the knowledge from normal events to detect anomalies. 
For example, if we fail to find "playing football on the road" in the knowledge about normality, we can assume that an anomalous event occurs.

\begin{figure}[!t]
	\centering
	\includegraphics[]{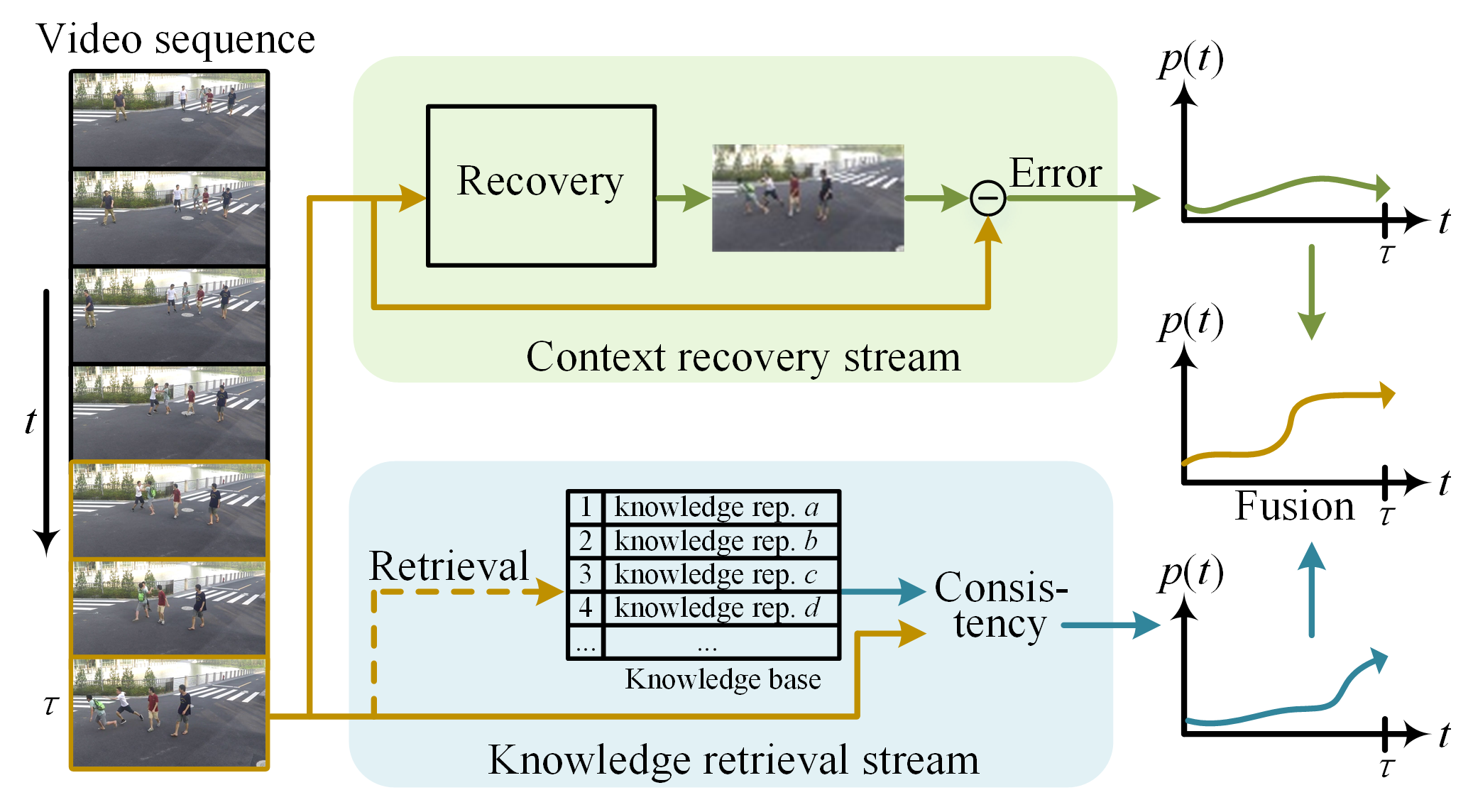}
	\caption{Overview of the proposed two-stream framework.
	The context recovery stream and the knowledge retrieval stream reflect the anomaly probability by the recovery error of the input and the consistency between the input and the knowledge representations about normality, respectively.
	The results from the two streams are fused as the final anomaly score.
}
	\label{fig1}
\end{figure}

In this work, we propose a novel two-stream framework that can not only discriminates short-term abnormal motions, but also leverages the knowledge from normal events to enhance the understanding of normality.
Different from other multi-stream models that use multiple modalities \cite{HierarchicalSpatioTemporal2021zeng, VariationalAbnormal2022li, FutureFrame2021luo, BackgroundAgnosticFramework2021georgescua}, our proposed two-stream framework consists of a context recovery stream and a knowledge retrieval stream, as shown in \figurename~\ref{fig1}.
In our context recovery stream, the normal short-term motion patterns are modeled by predicting the future frame based on the input snippet.
The anomaly probability of a testing event is obtained based on the prediction error.
In the knowledge retrieval stream, a series of normal events are encoded as the knowledge about normality and stored in a knowledge base.
The anomaly probability can be obtained according to the consistency between the testing event and the knowledge.
We fuse the anomaly probabilities from the two streams as the final anomaly score of the testing data.

For context recovery, the existing models \cite{FutureFrame2018liua, MemorizingNormality2019gonga, ClozeTest2020yu, LearningMemoryGuided2020parka, FutureFrame2021luo, LearningNormal2021lva, HybridVideo2021liua, AppearanceMotionMemory2021caia, RobustUnsupervised2021wanga, AnomalyDetection2022fang, VariationalAbnormal2022li} usually concatenate the input frames as an image and feed it into a 2D-CNN-based U-Net \cite{UnetConvolutional2015ronneberger} to reconstruct the snippet or predict the future frame.
To supplement the motion information, the constraint of optical flow is required.
Although there are a few works \cite{AnomalyDetection2021georgescua, DeepCascadeCascading2017sabokroua, DeepOneClass2020wu} adopt 3D CNN as the encoder, they do not explore a suitable 3D-CNN-based U-Net structure and can only use a shallow convolutional auto-encoder, which limits the representation capacity of the model.
We propose a novel spatiotemporal U-Net (STU-Net) for future frame prediction, which takes the 3D CNN designed for action recognition as the encoder and retains the temporal dimension in the deep layers, so as to extract rich semantic features of the motion.
To fuse the motion information, we add a temporal squeezing layer between the encoder and decoder, which also solves the inconsistency of temporal dimensions between the feature maps output by the encoder and those fed into the decoder.
In this way, we can take advantage of the motion information in the input frames to predict the future frame.
Besides, the existing context recovery method has the problem that the reconstruction or prediction error is proportional to the number of foreground objects \cite{SurveySingleScene2022ramachandra}, which is easy to cause false positives.
Some methods \cite{ObjectCentricAutoEncoders2019ionescua, ClozeTest2020yu, HybridVideo2021liua, AnomalyDetection2021georgescua, BackgroundAgnosticFramework2021georgescua} employ the object detector to solve this problem, but seriously ignore the scene-dependent characteristic of video anomalies.
Moreover, they are incapable of detecting the anomalous objects not included in the training classes of the object detector.
We propose a maximum local error (MLE) mechanism to focus on the recovery degree of the local anomalous region.
Specifically, based on the assumption that the anomalous region causes larger context recovery error than normal regions, we propose to use the maximum patch-level recovery error in the frame instead of the frame-level error to reflect the anomaly probability.
Due to the lack of validation sets in the existing datasets \cite{AnomalyDetection2010mahadevan, AbnormalEvent2013lua, RevisitSparse2017luoa, MultitimescaleTrajectory2020rodriguesa}, we use the training videos to generate pseudo anomalous samples by data augmentation to tune the hyper-parameter (\textit{i.e.} the size of the patch) in MLE.
The proposed MLE can partially ignore the recovery degree of the normal region, so that the recovery error of the anomalous region can be calculated accurately.
Meanwhile, it has the advantage of not relying on any object detectors.

For knowledge retrieval, although the representative features generated by some methods, such as the cluster centers in clustering \cite{ClusterAttention2020wanga} and the nearest neighbor samples in KNN \cite{StreetScene2020ramachandraa}, could be regarded as the knowledge about normality, they can hardly meet the requirements of efficiency, flexibility and comprehensiveness for knowledge extraction.
In this work, we propose an improved learnable locality-sensitive hashing (iL\textsuperscript{2}SH) based on \cite{LearnableLocalitySensitive2021lu} to store and retrieve the knowledge about normality, which can find the knowledge representation consistent with a testing event adaptively and efficiently.
Concretely, we first extract the features of the events in training videos and then encode them into hash codes via a trainable hash encoder composed of multiple parallel hash layers.
The binary and real-valued hash codes are stored as key-value pairs in multiple hash tables that serve as a knowledge base. 
We take the mean vector of the hash codes with the same key as the knowledge representation of such normal events.
A testing event obtains the hash code through the same process, and tries to find the knowledge representation sharing the same key.
We calculate the anomaly probability according to the distance between the testing hash code and the retrieved knowledge representation.
Compared with LLSH \cite{LearnableLocalitySensitive2021lu}, we make the following improvements.
First, we improve the optimization of the hash encoder.
LLSH adopts MoCo \cite{MomentumContrast2020he} contrastive learning framework to train the hash encoder, where the optimization effect is affected by the number of negative samples, and they need to set different numbers of negative samples for different datasets.
In contrast, we use a simpler Siamese network and discard negative samples, which can achieve better training results with only positive samples.
Second, the parameters in different hash layers should be as different as possible, which cannot be guaranteed in LLSH since it lacks constraints on the hash layers.
We propose a mutual difference loss to make the hash layers different from each other.
It can enlarge the distances between the hash layers and improve the performance after optimization.

We conduct comprehensive studies across several datasets to verify the effectiveness of the proposed two-stream framework, including ShanghaiTech \cite{RevisitSparse2017luoa}, CUHK Avenue \cite{AbnormalEvent2013lua}, IITB Corridor \cite{MultitimescaleTrajectory2020rodriguesa} and UCSD Ped2 \cite{AnomalyDetection2010mahadevan} datasets.
Without using optical flow and object detection, our context recovery stream can exceed the previous future frame prediction and snippet reconstruction methods \cite{FutureFrame2018liua, MemorizingNormality2019gonga, LearningMemoryGuided2020parka}.
Furthermore, the proposed iL\textsuperscript{2}SH outperforms other knowledge modeling methods, such as the clustering-based CAC \cite{ClusterAttention2020wanga} and the nearest-neighbor-search-based Exemplar Selection \cite{StreetScene2020ramachandraa}.
Through comparing with several existing models \cite{LearningMemoryGuided2020parka, LearningNormal2021lva, AppearanceMotionMemory2021caia, ClusterAttention2020wanga, StreetScene2020ramachandraa}, we verify that context recovery and knowledge retrieval can complement each other.
Our method achieves the state-of-the-art performance on all the four datasets.

We summarize our contributions as follows:
\begin{itemize}
	\item
	We propose a novel two-stream framework consisting of a context recovery stream and a knowledge retrieval stream for video anomaly detection.
	It not only utilizes the fine-level local context to detect anomalies, but also takes full advantage of the high-level semantic knowledge of normal events to enhance the understanding of normality.
	
	\item
	We propose a spatiotemporal U-Net and maximum local error mechanism to respectively enhance the ability of motion modeling of the auto-encoder and the ability of error calculation in anomalous regions, which significantly improve the accuracy of context recovery.

	\item
	We propose iL\textsuperscript{2}SH, which improves the optimization process of learnable locality-sensitive hashing.
	It can efficiently extract, store and retrieve the knowledge about normality, and detect anomalies according to the consistency between testing events and the knowledge.
	
	\item
	We prove that the context recovery stream and the knowledge retrieval stream are complementary for video anomaly detection by experiments.
	With the fusion of the two streams, our method achieves the state-of-the-art performance on ShanghaiTech, CUHK Avenue, IITB Corridor and UCSD Ped2 datasets.
	
\end{itemize}

\section{Related Work}
\subsection{Context Recovery Methods}
The context recovery methods \cite{FutureFrame2018liua, MemorizingNormality2019gonga, ClozeTest2020yu, BMANBidirectional2020lee, LearningMemoryGuided2020parka, FutureFrame2021luo, SmithNetStrictness2021nguyen, LearningNormal2021lva, HybridVideo2021liua, AppearanceMotionMemory2021caia, RobustUnsupervised2021wanga, MultiEncoderEffective2021fang, AnomalyDetection2022fang, VariationalAbnormal2022li, SelfSupervisedAttentive2022huang, InfluenceawareAttention2022zhang}, which are based on the assumption that normal events are easy to recover while the abnormal events are hard to recover, become the mainstream in the field of VAD in recent years.
This kind of methods reconstruct the snippet or predict the future frame through an encoder-decoder-style generative model.
For example, \cite{FutureFrame2018liua, FutureFrame2021luo, LearningMemoryGuided2020parka, LearningNormal2021lva, AppearanceMotionMemory2021caia, VariationalAbnormal2022li} concatenate the input frames as an image, and feed it into a U-Net model to predict the next frame.
However, they directly apply a 2D U-Net proposed in the field of image segmentation to videos, making it difficult to model the motion of objects.
Hence, they have to combine with the optical flow modality \cite{FutureFrame2018liua, FutureFrame2021luo, AppearanceMotionMemory2021caia} or recurrent units \cite{BMANBidirectional2020lee, RevisitSparse2017luoa}.
For example, Lee \textit{et al.} \cite{BMANBidirectional2020lee} adopt ConvLSTM to encode the spatiotemporal features in both forward and backward directions.
There are some methods that adopt a 3D CNN as the encoder \cite{DeepCascadeCascading2017sabokroua, AnomalyDetection2021georgescua, DeepOneClass2020wu}.
Nevertheless, these models usually use shallow networks to avoid gradient vanishing.
Compared with the above methods, our proposed STU-Net can model the motion information without additional modality or recurrent operation.
There is a 3D-CNN-based U-Net in the field of image summary \cite{VideoSummarization2022liu}.
It directly applies squeeze-and-excitation blocks \cite{SqueezeandExcitationNetworks2020hu} for compressing feature dimensionality and thus cannot be applied for future frame prediction, in which the encoder and decoder have different numbers of frames.
In contrast, we propose temporal squeezing layers in our STU-Net and solve the problem of inconsistent temporal dimensions between the feature maps of the encoder and the decoder.

Besides, most of the methods feed the whole frame into the model in both training and testing phases, and others take the detected objects \cite{HybridVideo2021liua, BackgroundAgnosticFramework2021georgescua, ClozeTest2020yu, AnomalyDetection2021georgescua, ObjectCentricAutoEncoders2019ionescua} or video patches \cite{DeepCascadeCascading2017sabokroua, StreetScene2020ramachandraa, LearningDistance2020ramachandraa} as the input.
As described in the Introduction section, the methods based on object detection only use the foreground objects and thus ignore the scene information.
Additionally, they cannot detect the anomalous objects whose categories are not covered in the pre-trained object detector.
The patch-based methods are difficult to capture the full movement of the object, since patches of the same spatial grid across times incurs visual misalignment when rigidly dividing a moving object.
In our context recovery stream, we still use the whole frame as the input so that complete and long-term motions can be captured.
In the testing phase, we adopt the maximum error among the patches of the recovery error map (\textit{i.e.} a frame) to reveal the anomaly score.
Although Nguyen and Meunier \cite{AnomalyDetection2019nguyena} also train a frame-level model and calculate normalized patch-level errors, they do not exploit an effective solution to solve the large error problem caused by foreground objects.
They use a fixed and small patch for different datasets, which is easily affected by the noise in the error map and cannot handle different resolutions of the videos.
We propose a novel maximum local error (MLE) mechanism that utilizes the training videos to simulate anomalies and selects an appropriate patch size for the dataset.
The proposed MLE can effectively focus on the anomalous region and calculate a more accurate recovery error.
Therefore, it alleviates the problem of recovery error proportional to the number of foreground objects.

\subsection{Knowledge Retrieval Methods}
The knowledge retrieval methods explicitly extract the representations of training data as the knowledge about normality, and detect anomalies according to the consistency between the testing event and the knowledge representations.
The commonly used knowledge representations include decision boundaries of OC-SVM \cite{DeepAppearance2017smeureanu, ObjectCentricAutoEncoders2019ionescua, LearningDeep2015xua}, nearest neighbor samples \cite{VideoAnomaly2012saligramaa, StreetScene2020ramachandraa}, cluster centers \cite{DetectingAbnormal2019ionescua, ClusterAttention2020wanga, GraphEmbedded2020markovitza, ClusteringDriven2020chang} and probability distributions \cite{GaussianProcess2015cheng, VideoAnomaly2020fana}.
For example, Ionescu \textit{et al.} \cite{ObjectCentricAutoEncoders2019ionescua} first classify the normal data into \textit{K} classes by K-means, and then use the maximum classification score from \textit{K} one-versus-rest SVMs as the anomaly score.
Wang \textit{et al.} \cite{ClusterAttention2020wanga} propose a cluster attention module to map the input event into \textit{K} feature spaces.
For a testing sample, the highest similarity between its \textit{K} feature space representations and corresponding space centers is regarded as its regularity score.
However, the value of \textit{K} is usually too small to fully exploit the knowledge in normal data.
Ramachandra and Jones \cite{StreetScene2020ramachandraa} build an exemplar set, in which only the normal sample whose distance from the stored samples in the set exceeds a threshold will be added.
The anomaly score is determined simply based on the distance between the testing sample and its nearest exemplars, which lacks the abstraction of knowledge.
The normal patterns in the memory modules of context recovery methods \cite{MemorizingNormality2019gonga, LearningMemoryGuided2020parka, LearningNormal2021lva, HybridVideo2021liua, AppearanceMotionMemory2021caia} can also be regarded as the knowledge representations.
Limited by the size of the memory, it is difficult to contain adequate knowledge in the memory module.
Compared with the above methods, our proposed iL\textsuperscript{2}SH can make use of the knowledge from training data and retrieve it efficiently.
Even if iL\textsuperscript{2}SH is combined with the memory-augmented context recovery model, it is still able to bring significant improvements.
Different from Lu \textit{et al.} \cite{LearnableLocalitySensitive2021lu} who use MoCo \cite{MomentumContrast2020he} to train the hash functions in their hash encoder, which is easily affected by the number of negative samples (\textit{i.e.} the length of queue in MoCo), we use Siamese network for optimization and discard negative samples. In addition, we propose a new loss which can enlarge the differences between any two hash functions.

\begin{figure}[!t]
	\centering
	\includegraphics[]{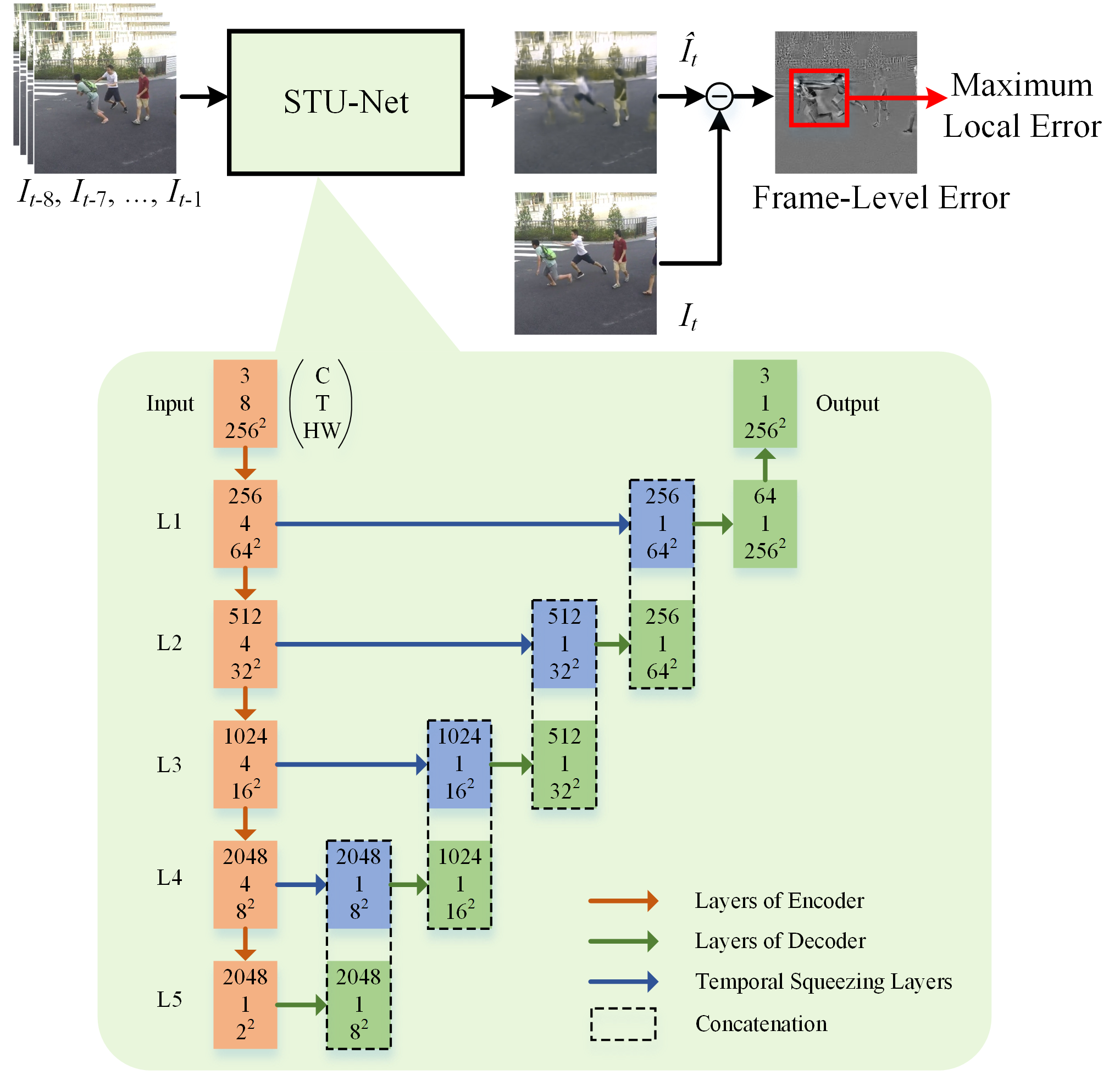}
	\caption{Overview of our context recovery stream. A clip of previous 8 frames are fed into STU-Net and predict the current frame. The maximum local error (in red box) between the predicted frame and the ground truth frame is used as the anomaly score.}
	\label{fig2}
\end{figure}

\section{Proposed Method}
In this section, we first introduce the architecture of the proposed two-stream framework.
Then, we illustrate the spatiotemporal U-Net (STU-Net) in our context recovery stream, followed by the introduction of maximum local error (MLE) mechanism.
Next, we describe the proposed improved learnable locality-sensitive hashing (iL\textsuperscript{2}SH) in the knowledge retrieval stream, which includes the sequential processes of training hash encoder, constructing knowledge base and retrieving knowledge.
Finally, we introduce the fusion of anomaly scores of the two streams.

\subsection{Two-Stream Framework}
The proposed two-stream framework for video anomaly detection is shown in \figurename~\ref{fig1}.
To detect if an anomalous event occurs in a video sequence at time $\tau$, the snippet containing the $\tau$-th frame is fed into the context recovery stream and the knowledge retrieval stream, respectively.
In the context recovery stream, the input is recovered by an encoder-decoder-style future frame prediction model in our implementation.
The error between the recovered frame and the ground truth frame is taken as the anomaly score.
In the knowledge retrieval stream, the knowledge base contains the knowledge about normality extracted from training data.
The normal knowledge representation consistent with the input event is retrieved from the knowledge base, and the anomaly score is determined by the distance between the knowledge representation and the event.
If no knowledge representation can be retrieved, the anomaly score is given a high value.
The anomaly scores of the two streams are added together as the final anomaly score at time $\tau$.

\subsection{Context Recovery Stream}
We propose a novel future frame prediction model named spatiotemporal U-Net (STU-Net) for the context recovery stream.
STU-Net takes a snippet as the input and predicts the next frame, as illustrated in \figurename~\ref{fig2}.
In order to detect the anomaly for the current frame $I_t$, the previous 8 frames $I_{t-8}, I_{t-7}, \cdots, I_{t-1}$ are fed into STU-Net to generate the predicted frame $\hat{I}_t$. 
We calculate the proposed maximum local error (MLE) between $\hat{I}_t$ and its ground truth $I_t$ as the anomaly score.

\subsubsection{Spatiotemporal U-Net}
\begin{table}[!t]
	\centering
	\caption{Network Details of Spatiotemporal U-Net}
	\label{tab1}
	\setlength\tabcolsep{0.9pt}
	\begin{tabular}{@{}cclcccc@{}}
		\toprule
		\multicolumn{7}{c}{Spatiotemporal U-Net}                                                                                                                           \\ \midrule
		\multicolumn{4}{c}{Encoder}                                                                      & \multicolumn{3}{c}{Decoder}                                             \\ \midrule
		\textit{fn1}(\textit{n}):             & \multicolumn{3}{c}{{[}(\textit{n},3,1\textsuperscript{2}), (\textit{n},1,3\textsuperscript{2}),   (4\textit{n},1,1\textsuperscript{2}){]}}                  & \multirow{2}{*}{\textit{fn3}(\textit{n}):} & \multicolumn{2}{c}{{[}(\textit{n},1,3\textsuperscript{2}), (\textit{n},1,3\textsuperscript{2}),}~~~   \\ \cmidrule(lr){1-4}
		\textit{fn2}(\textit{n}):             & \multicolumn{3}{c}{{[}(\textit{n},1,1\textsuperscript{2}), (\textit{n},1,3\textsuperscript{2}),   (4\textit{n},1,1\textsuperscript{2}){]}}                  &                          & \multicolumn{2}{c}{(4\textit{n},1,3\textsuperscript{2}), (2\textit{n},1,2\textsuperscript{2})\textsuperscript{$\mathsf{T}$}{]}} \\ \midrule
		Level               & Kernel        & \multicolumn{1}{c}{Stride} & Shape                        & Level                    & Kernel        & Shape                       \\ \midrule
		Input               & -             & \multicolumn{1}{c}{-}      & (3,8,256\textsuperscript{2})                    & Input                    & -             & (2048,1,2\textsuperscript{2})                  \\ \midrule
		\multirow{4}{*}{L1} & (64,5,72)     & (1,2\textsuperscript{2})                     & (64,8,128\textsuperscript{2})                   & \multirow{4}{*}{L5}      & (512,1,1\textsuperscript{2})    & \multirow{2}{*}{(512,1,4\textsuperscript{2})}  \\ \cmidrule(lr){2-4}
		& \textit{max}(1,3\textsuperscript{2})     & (1,2\textsuperscript{2})                     & (64,8,64\textsuperscript{2})                    &                          & (512,1,2\textsuperscript{2})\textsuperscript{$\mathsf{T}$}   &                              \\ \cmidrule(lr){2-4} \cmidrule(l){6-7} 
		& \textit{fn1}(64)$\times$3     & (1,1\textsuperscript{2})$\times$9                   & (256,8,64\textsuperscript{2})                   &                          & (512,1,1\textsuperscript{2})    & \multirow{2}{*}{(2048,1,8\textsuperscript{2})} \\ \cmidrule(lr){2-4}
		& \textit{max}(2,1\textsuperscript{2})     & (2,1\textsuperscript{2})                     & (256,4,64\textsuperscript{2})                   &                          & (2048,1,1\textsuperscript{2})\textsuperscript{$\mathsf{T}$}  &                              \\ \midrule
		\multirow{3}{*}{L2} & {[}\textit{fn1}(128),  & (1,1\textsuperscript{2})                     & \multirow{3}{*}{(512,4,32\textsuperscript{2})}  & \multirow{2}{*}{L4}                       & \textit{fn3}(512)      & \multirow{2}{*}{(1024,1,16\textsuperscript{2})}                 \\
		& \textit{fn2}(128){]}$\times$2 & (1,2\textsuperscript{2})                     &                               &                          &               &                              \\ \cmidrule(l){5-7} 
		&               & (1,1\textsuperscript{2})$\times$10                  &                               & \multirow{2}{*}{L3}                       & \textit{fn3}(256)      & \multirow{2}{*}{(512,1,32\textsuperscript{2})}                  \\ \cmidrule(r){1-4}
		\multirow{3}{*}{L3} & {[}\textit{fn1}(256),  & (1,1\textsuperscript{2})                     & \multirow{3}{*}{(1024,4,16\textsuperscript{2})} &                          &               &                              \\ \cmidrule(l){5-7} 
		& \textit{fn2}(256){]}$\times$3 & (1,2\textsuperscript{2})                     &                               & \multirow{2}{*}{L2}                       & \textit{fn3}(128)      & \multirow{2}{*}{(256,1,64\textsuperscript{2})}                  \\
		&               & (1,1\textsuperscript{2})$\times$16                  &                               &                          &               &                              \\ \midrule
		\multirow{3}{*}{L4} & \textit{fn2}(512)      & (1,1\textsuperscript{2})                     & \multirow{3}{*}{(2048,4,8\textsuperscript{2})}  & \multirow{2}{*}{L1}      & \textit{fn3}(64)       & (128,1,128\textsuperscript{2})                 \\ \cmidrule(l){6-7} 
		& \textit{fn1}(512)      & (1,2\textsuperscript{2})                     &                               &                          & \textit{fn3}(32)       & (64,1,256\textsuperscript{2})                  \\ \cmidrule(l){5-7} 
		& \textit{fn2}(512)      & (1,1\textsuperscript{2})$\times$7                   &                               &\multirow{2}{*}{Output}                          & (64,1,3\textsuperscript{2})     & \multirow{2}{*}{(3,1,256\textsuperscript{2})}  \\ \cmidrule(r){1-4}
		L5                  & \textit{avg}(4,7\textsuperscript{2})     & (1,1\textsuperscript{2})                     & (2048,1,2\textsuperscript{2})                   &                          & (3,1,3\textsuperscript{2})$\times$2    &                              \\ \bottomrule
	\end{tabular}
\end{table}

As shown in \figurename~\ref{fig2}, the proposed STU-Net is a 5-level encoder-decoder with U-Net architecture.
A clip of 8 frames with the spatial resolution of 256$\times$256 are input into it.
The encoder gradually reduces the spatial and temporal resolutions of the input to extract high-level semantic features, while the decoder gradually recovers the feature map by increasing the spatial resolution.
To avoid the gradient vanishing problem, the feature maps of the encoder and decoder in each level are connected via a shortcut.

Different from the previous works that stack the input frames as an image to use 2D convolutions, we retain the temporal dimension in the encoder, which helps to predict the future frame through the motion in previous frames.
Due to the inequality in temporal dimension, the feature map output by the encoder cannot be directly connected to the input of the decoder in the same level.
We address this issue by adding a temporal squeezing layer (TSL) on each shortcut.
The TSL fuses the features from different time steps and squeezes the temporal dimension to 1.
In this way, the feature maps from the encoder and the decoder in the same level can be concatenated along the channel dimension, to serve as the input of the next level of the decoder.

The network structures of the encoder and decoder are detailedly displayed in \tablename~\ref{tab1}.
We adopt the I3D network \cite{QuoVadis2017carreira} with ResNet-50 backbone \cite{DeepResidual2016he} as the encoder, and design the decoder on our own.
Each parenthesis in the table describes the kernel size of the convolution/pooling layer or the shape of the output feature map in dimension $(C, T, HW)$, where $C$, $T$, $H$ and $W$ respectively represent the number of channels, the length, height and width of the feature map.
For strides and pooling layers, we omit the channel dimension $C$.
We define three functions $fn\mathit{1}$, $fn\mathit{2}$ and $fn\mathit{3}$ for convenience. Each of them represents a block composed of multiple convolution layers.
$max$ and $avg$ denote the max pooling and average pooling, respectively.
The residual connection in each level is not displayed.

For the layers in the encoder, each convolution is followed by a batch normalization (BN) \cite{BatchNormalization2015ioffe} and a ReLU activation.
We remove the last ReLU activation, as it restricts diverse feature representations.
In the decoder, the convolution layer with superscript $\mathsf{T}$ represents a transposed convolution, which enlarges the spatial resolution of the feature map.
The strides in regular convolutions and transposed convolutions are $(1, 1^2)$ and $(1, 2^2)$, respectively.
Except for the last layer, each convolution is followed by a BN and a Leaky ReLU.
The temporal squeezing layers are not shown in the table. There is only a convolution with kernel size $(C, 4, 1^2)$ and stride $(1, 1^2)$ in each TSL.

We minimize the mean square error (MSE) and L1 loss between the predicted frame $\hat{I}_t$ and the ground truth $I_t$ for training STU-Net:
\begin{equation}
	\hat{I}_t = ~ \mathrm{STUNet}([I_{t-8}, I_{t-7}, \cdots, I_{t-1}]),
\end{equation}
\begin{equation}
	L_{FLE}(I_t, \hat{I}_t) = ~ \Vert I_t - \hat{I}_t \Vert _{F}^{2} + \lambda_{L1} \times \vert I_t -\hat{I}_t \vert,
	\label{eq2}
\end{equation}
where $\lambda_{L1}$ is the weight of L1 loss, and $L_{FLE}$ denotes the loss of frame-level error (FLE).

The proposed STU-Net has the ability to utilize the temporal information of the input snippet to predict the future frame.
By using the frame-level error (\textit{i.e.} Eq. (\ref{eq2})) as anomaly score, it is comparable to existing frame prediction methods that combine with optical flow to supplement motion information.

\subsubsection{Maximum Local Error}
\begin{algorithm}[!t]
	\caption{The Maximum Local Error Mechanism}
	\begin{algorithmic}
		\Require $V=\{I_i\}_{i=1}^{N}$: a training video of $N$ frames \\
		\quad $nseg$: the number of anomalous segments in a video (default: 1) \\
		\quad $ratio \in (0, 1) $: the ratio of anomaly frames in each segment (default: 0.5) \\
		\quad $offset$: the offset index of the frame to be averaged with current frame (default: 2)
		\Ensure an anomalous video $\tilde{V}=\{\tilde{I}_i\}_{i=1}^{N}$, and a list of labels $\{L_i\}_{i=1}^{N}$ indicating if the frame is normal (0) or not (1)
		\State \textbf{function} rotate($I$): rotate $I$ with a random angle $\alpha \in [\ang{2}, \ang{5}]$
		\State \textbf{function} flip($I$): horizontally flip $I$
		\State $m \gets \mathrm{floor}(N / nseg) $ \Comment{get the length of a segment}
		\For{each $i \in \{1, 2, \cdots, N\}$}
			\State $j \gets i \mod m $ \Comment{get the segment index}
			\State $start \gets \mathrm{floor}(m \times (j + 0.5 - 0.5 \times ratio )) + 1$
			\State $end \gets start + \mathrm{floor}(m \times ratio)$
			\State $\tilde{I}_i \gets \mathrm{rotate}(\mathrm{flip}(I_i))$
			\If{$i >= start$ and $i < end$}
				\State $\tilde{I}_{f} \gets \mathrm{rotate}(\mathrm{flip}(I_{i+offset}))$
				\State $\tilde{I}_i \gets 0.5 \times (\tilde{I}_i + \tilde{I}_{f})$
				\State $L_i \gets 1$
			\Else
				\State $L_i \gets 0$
			\EndIf
		\EndFor

	\end{algorithmic} 
	\label{alg1}
\end{algorithm}

For the VAD task, we expect the context recovery model to be accurate in predicting normal regions and inaccurate in abnormal regions.
However, because of the nonstatic background, large number of foreground objects, image noise and other possible factors, we cannot make fully accurate predictions for normal regions in the next frame in any case.
Inaccurate prediction of normal regions will lead to higher errors and false positives.

To pay more attention to the anomalous region and ignore normal regions, we propose a maximum local error (MLE) for anomaly detection process, as shown in the red box in \figurename~\ref{fig2}.
We use a square sliding window with fixed size to calculate a number of local errors on the frame-level error map.
Based on the hypothesis that the error of anomalous region is larger than that of normal region, we choose the maximum local error as the anomaly score.
The proposed MLE can be implemented by max pooling.
Mathematically, MLE is denoted as:
\begin{align}
	\begin{split}
			MLE(I_t, \hat{I}_t)_{k,s} = \mathrm{MaxPool}_{k,s}(\Vert I_t - \hat{I}_t \Vert _{F}^{2} + \\
			 \lambda_{L1} \times \vert I_t -\hat{I}_t \vert),
		\end{split}
\end{align}
where $k$ is the size of the sliding window, and $s$ is the stride.

The hyper-parameter $k$ is dataset-based and can be determined by a validation set.
However, most VAD datasets do not have validation sets.
Thus, we propose an alternative solution that uses training videos to simulate anomalies by means of data augmentation, which can be seen in Algorithm \ref{alg1}.
Specifically, we spatially flip and rotate all the frames in a video at a random angle to obtain a new video that the model has not seen before.
To generate an anomalous frame, we fuse the current frame with its future frame by averaging.
We use the simulated abnormal videos for video anomaly detection, and select an appropriate $k$ from the predefined set $K=\{k_1, k_2, \cdots, k_n\}$ according to the evaluation metric.

The proposed MLE mitigates the interference of recovery errors in normal regions.
With MLE, our STU-Net can reach similar or better performance compared with the methods using object detectors.

\subsection{Knowledge Retrieval Stream}
\begin{figure}[!t]
	\centering
	\includegraphics[]{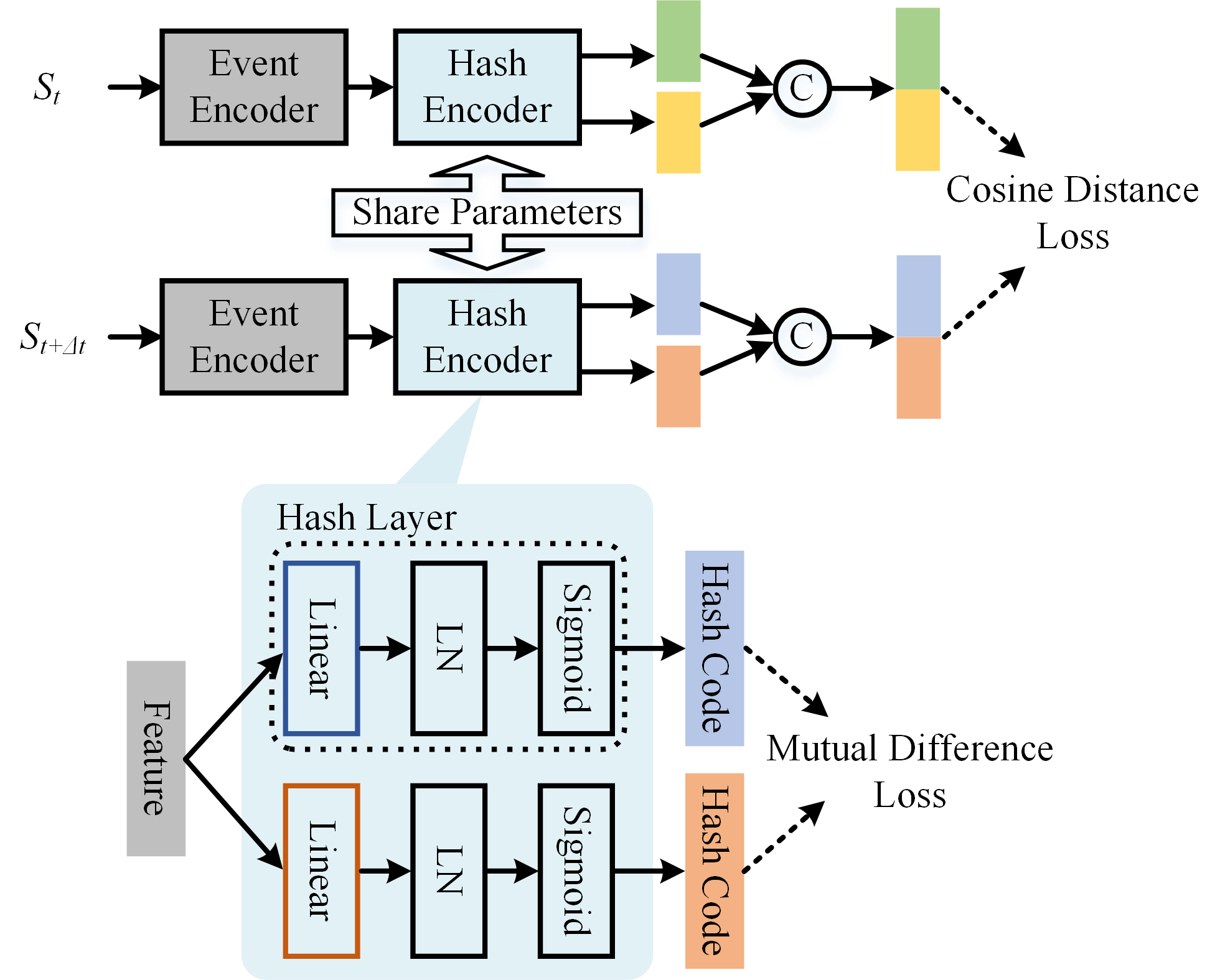}
	\caption{The architecture and training process of iL\textsuperscript{2}SH. The proposed iL\textsuperscript{2}SH includes an event encoder and a hash encoder which consists of a group of hash layers. We use a Siamese network for training iL\textsuperscript{2}SH, where the two branches share the same parameters.}
	\label{fig3}
\end{figure}
The knowledge retrieval stream is proposed for enhancing the understanding of normality.
We aim to adaptively construct a knowledge base, store the knowledge about normality from training data, and retrieve the knowledge efficiently to detect anomalies.
To this end, we propose an improved learnable locality-sensitive hashing (iL\textsuperscript{2}SH).
First, we take hash functions as learnable parameters and embed them into a neural network to optimize using the training data.
Then, we map all training events into hash codes by the optimized hash functions.
The mean vector of similar hash codes are stored in each bucket of the hash table and served as the normal knowledge representations.
Finally, in the testing phase, we look up a bucket consistent with the testing event, and calculate the anomaly score based on the distances between the testing hash codes and the retrieved knowledge representation.

\subsubsection{Training iL\textsuperscript{2}SH}
The structure of iL\textsuperscript{2}SH is illustrated in \figurename~\ref{fig3}, which mainly consists of an event encoder and a hash encoder.
The event encoder outputs a feature vector to represent the event in the input snippet.
In this work, it is the I3D network as introduced in the context recovery stream.
The hash encoder contains a group of parallel hash layers, each of which maps the feature to a real-valued hash code and will be used to construct a hash table.
A hash layer has three sequential layers including a liner layer, a layer normalization \cite{LayerNormalization2016ba} and a sigmoid activation.
Each linear layer serves as a hash function and we aim to optimize it to generate similar hash codes for similar features.

\begin{algorithm}[!t]
	\caption{The Process of Constructing Knowledge Base} 
	\begin{algorithmic}
		\Require $\{S_i\}_{i=1}^{N}$: $N$ training snippets \\
		\quad $Enc(\cdot)$: iL\textsuperscript{2}SH, which maps a snippet to $B$ hash codes \\
		\quad $\{\mathcal{H}_b[key]=(cnt, val)\}_{b=1}^{B}$: $B$ empty hash tables
		\Ensure hash tables $\{\mathcal{H}_b\}_{b=1}^{B}$ that stores the hash codes
		\Function{bin}{$a$} \Comment{Binary function}
			\ForAll{$a^{<i>}$ of the $i$-th bit in $a$}
				\State $a^{<i>}$ $\gets$ $0$ if {$a^{<i>} < 0.5 $} else $1$
			\EndFor
			\State \Return $a$
		\EndFunction
		
		\For{each $i \in \{1,2,\cdots, N\}$}
			\State $\{h_{b}\}_{b=1}^{B}$ $\gets$ $Enc(S_i)$ 
			\For{each $b \in \{1, 2, \cdots, B\}$}
				\State $k$ $\gets$ \Call{bin}{$h_b$}
				\If{$k$ exists in $\mathcal{H}_b.key$}
					\State $\mathcal{H}_b[k].val \gets (\mathcal{H}_b[k].val \times \mathcal{H}_b[k].cnt + h_{b})~/~{(\mathcal{H}_b[k].cnt + 1)}$
					\State $\mathcal{H}_b[k].cnt \gets \mathcal{H}_b[k].cnt + 1$
				\Else
					\State $\mathcal{H}_b[k].cnt \gets 1$
					\State $\mathcal{H}_b[k].val \gets h_{b}$
				\EndIf
			\EndFor
		\EndFor
		
	\end{algorithmic} 
	\label{alg2} 
\end{algorithm}

As shown in \figurename~\ref{fig3}, iL\textsuperscript{2}SH is embedded as a branch of the Siamese network, where the two branches share the same parameters.
We feed a snippet $S_t$ into one of the branches.
At the same time, a similar snippet $S_{t+\varDelta t}$ which is temporally close to $S_t$ is sampled from the same video and fed into the other branch.
Each branch outputs a group of short hash codes, which are concatenated as a compact long hash code.
We minimize the cosine distance between the concatenated hash codes of the two branches:
\begin{equation}
	L_{c}(l_t, l_{t+\varDelta t}) = 1 - \frac{l_t}{\Vert l_t\Vert} \cdot \frac{l_{t+\varDelta t}}{\Vert l_{t+\varDelta t}\Vert}
	\label{eq4}
\end{equation}
where $l_s$ and $l_{t+\varDelta t}$ is the concatenated hash codes corresponding to the input snippets $S_t$ and $S_{t+\varDelta t}$.

$S_t$ and $S_{t+\varDelta t}$ make up a positive pair and the distance between them is pulled close by Eq. (\ref{eq4}).
We do not take the snippets from different videos as negative pairs and push away the distances between them, since they provide little improvement for training and the number of negative pairs is sensitive to the scale of dataset.
Instead, we expect different hash layers to output as different hash codes as possible to construct different hash tables.
Therefore, we propose a mutual difference loss that enlarges the difference between the hash codes output by a hash encoder:
\begin{equation}
	L_{m}([h_1, h_2, \cdots, h_B]) = \frac{2}{RB(B-1)} \sum_{j=1}^{B} \sum_{i=j+1}^{B} \frac{h_i}{\Vert h_i \Vert} \cdot \frac{h_j}{\Vert h_j \Vert},
\end{equation}
where $h_i \in \mathbb{R}^{R}$ is a hash code of length $R$, and $B$ denotes the number of hash codes.

We average the mutual difference losses of the two branches.
The total loss for training iL\textsuperscript{2}SH is denoted as:
\begin{equation}
	L_{total} = L_c + \frac{\lambda_m}{2} (L_m^{(1)} + L_m^{(2)}),
\end{equation}
where $L_m^{(i)} (i\in\{1, 2\})$ denotes the mutual difference loss of the $i$-th branch, and $\lambda_m$ is the weight.

When the training process is finished, the hash layers has an enhanced ability to map similar events to similar hash codes.
We can use iL\textsuperscript{2}SH for constructing knowledge base in the next step.

\subsubsection{Constructing Knowledge Base}
\begin{figure}[!t]
	\centering
	\includegraphics[]{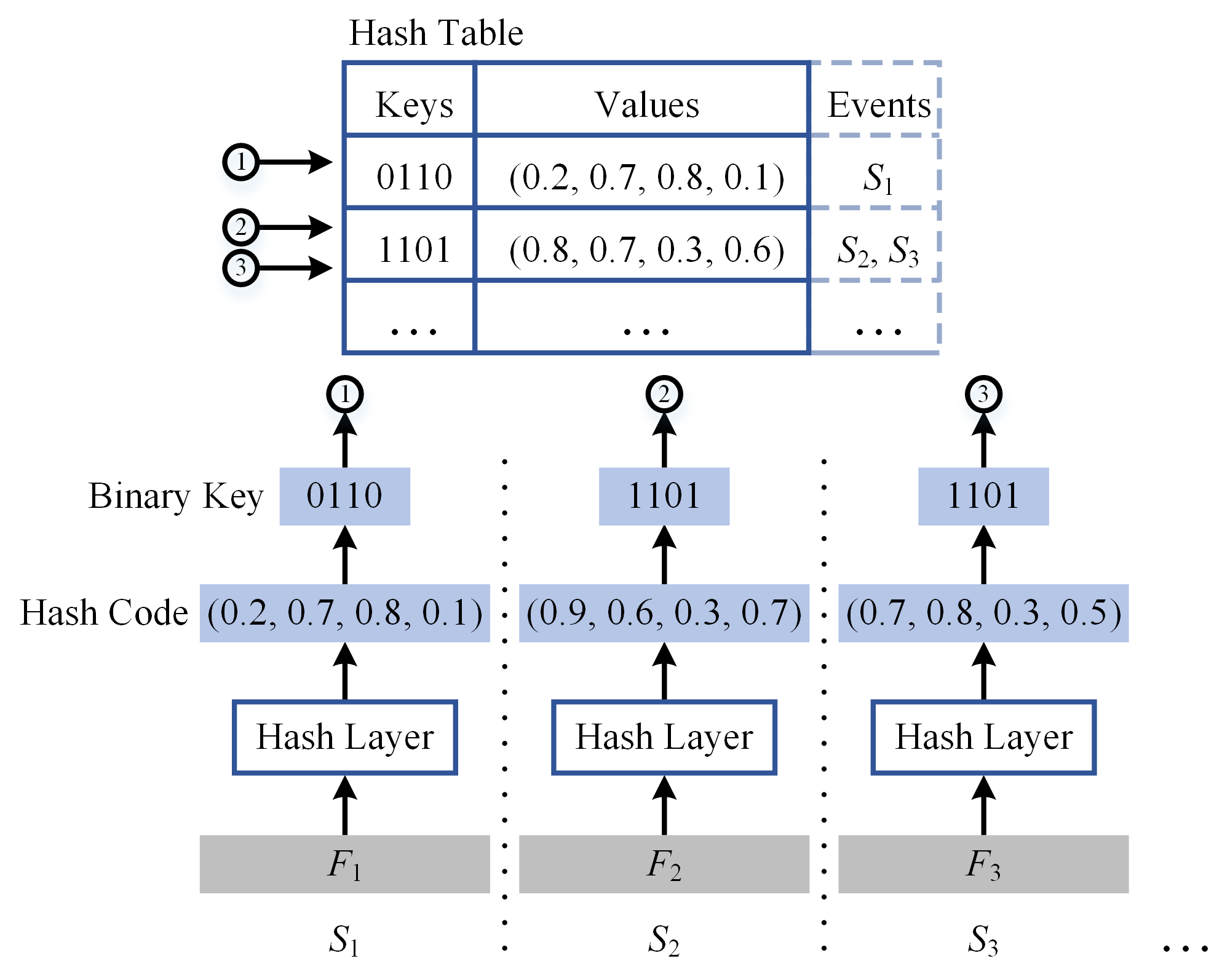}
	\caption{An example of constructing a hash table. $F$ denotes a feature fed into the hash layer. The hash codes with the same key are averaged and stored in one bucket.}
	\label{fig4}
\end{figure}

Our purpose is to construct a knowledge base which contains the knowledge representations about normality obtained from the training data.
To this end, we first map each training event to a group of hash codes, and then store each hash code into a corresponding hash table.
Each hash table is composed of a number of buckets in the form of key-value pairs.
The keys are the binary vectors of the hash codes, and the values are the mean vectors of those hash codes sharing the same binary key.
A detailed process of constructing knowledge base is summarized in Algorithm \ref{alg2}.

We take one hash layer as an example to explain the process of constructing a hash table, which is shown in \figurename~\ref{fig4}.
The normal events $S_1$, $S_2$ and $S_3$ generate three real-valued hash codes via the same hash layer.
Then we use the binary function in Algorithm \ref{alg2} to obtain a binary key for each hash code.
The hash code of $S_1$ is stored in the first bucket with its binary key "0110".
Since $S_2$ and $S_3$ has the same binary key "1101", we calculate the mean vector of the two hash codes, which is then stored in the second bucket.
In this way, similar events are abstracted into a knowledge representation and stored as a vector in a bucket.
Meanwhile, each knowledge representation can be retrieved efficiently via the binary key, which will be introduced in the following step.

\subsubsection{Retrieving Knowledge}
\begin{algorithm}[!t]
	\caption{The Process of Retrieving Knowledge}
	\begin{algorithmic}
		\Require $S_t$: a testing snippet \\
		\quad $Enc(\cdot)$: iL\textsuperscript{2}SH, which maps a snippet to $B$ hash codes \\
		\quad $\{\mathcal{H}_b[key]=(cnt, val)\}_{b=1}^{B}$: $B$ hash tables obtained by Algorithm \ref{alg2} \\
		\quad $P_{max}$: a predefined maximum anomaly score
		\Ensure $p_t$: the anomaly score of $S_t$

		\State $\{h_{b}\}_{b=1}^{B}$ $\gets$ $Enc(S_t)$ 
		\State $p_t$ = $P_{max}$
		\For{each $b \in \{1, 2, \cdots, B\}$}
			\State $k$ $\gets$ \Call{bin}{$h_b$} \Comment{\Call{bin}{} is the function in Algorithm \ref{alg2}}
			\If{$k$ exists in $\mathcal{H}_b.key$}
				\State $d$ = $\Vert \mathcal{H}_b[k].val - h_b \Vert_2$
				\If {$d < p_t$}
					\State $p_t \gets d$
				\EndIf
			\EndIf
		\EndFor

	\end{algorithmic} 
	\label{alg3} 
\end{algorithm}

Given a testing snippet, we aim to discriminate whether it is consistent with the normal knowledge and estimate an anomaly probability.
Therefore, we try to retrieve a knowledge representation from the knowledge base and use the distance between it and the retrieved knowledge representation as the anomaly score.
Algorithm \ref{alg3} describes the process of retrieving knowledge.
For a testing snippet $S_t$, we first obtain a group of hash codes and binary keys.
Then, we retrieve a bucket from the corresponding hash table, and calculate the L2 distance between the testing hash code and the vector in the retrieved bucket.
The minimum distance from all the hash tables is taken as the anomaly score of $S_t$.
By using the decision from multiple hash tables, we can find the most relevant knowledge representation and hence reduce false alerts.
However, each of the binary keys may not exist in the corresponding hash table.
In this case, we treat $S_t$ as an anomalous event which is inconsistent with normal knowledge.
$S_t$ is assigned with a predefined high anomaly score $P_{max}$, which is the maximum L2 distance between any two hash codes:
\begin{equation}
	P_{max} = \sqrt{R},
\end{equation}
where $R$ is the length of a hash code.

In summary, we construct a group of hash tables as the knowledge base, retrieve a knowledge representation (\textit{i.e.} mean vector of hash codes) from each hash table, and compare the testing event with the retrieved knowledge representations to detect anomalies.

\subsection{Fusion of Two Streams}
Now we can obtain the anomaly scores from the context recovery stream and the knowledge retrieval stream.
We fuse the results from the two streams by a late fusion:
\begin{equation}
	p_{fuse} = \lambda_{cr} p_{cr} + \lambda_{kr} p_{kr},
\end{equation}
where $p_{cr}$ and $p_{kr}$ respectively denote the anomaly score obtained from the context recovery stream and the knowledge retrieval stream, and $\lambda_{cr} > 0$ and $\lambda_{kr} > 0$ are the corresponding weights.

The context recovery stream has a good ability to detect short-term anomalous movements, and the knowledge retrieval stream can make use of high-level semantic knowledge about normality to detect anomalous events.
The anomaly detection results from the two streams can complement each other and thereby improve the performance for VAD.

\section{Experiments}
In this section, we conduct comprehensive experiments to verify the effectiveness of our proposed two-stream framework and compare with other methods.
We first introduce the datasets, evaluation metric and implementation details in our experiments.
Next, we carry out a detailed ablation study to investigate the effect of different components proposed in our method.
Then, we study the complementarity of context recovery and knowledge retrieval by fusing different types of existing methods.
After that, we visualize and analyze the effect of MLE and fusion of two streams.
Finally, we compare our two-stream framework with existing methods, where our method achieves the state-of-the-art performance.

\subsection{Datasets}
\begin{table*}[!t]
	\centering
	\caption{Four Datasets Used in Our Experiments}
	\label{tab2}
	\begin{tabular}{@{}clccccl@{}}
		\toprule
		\multicolumn{1}{l}{Dataset} & Year & \begin{tabular}[c]{@{}c@{}}Training \\ videos / frames\end{tabular} & \begin{tabular}[c]{@{}c@{}}Testing \\ videos / frames\end{tabular} & \multicolumn{1}{l}{Scenes} & \multicolumn{1}{l}{Resolution} & Abnormal events                                    \\ \midrule
		UCSD Ped2 \cite{AnomalyDetection2010mahadevan}           	& 2010 & 16 / 2.6k        & 12 / 2.0k      & 1   & 360$\times$240   & Bikers, carts and skaters                               \\
		CUHK Avenue \cite{AbnormalEvent2013lua}         			& 2013 & 16 / 15k         & 21 / 15k       & 1   & 640$\times$360   & Throwing object, wrong direction, running, \textit{etc.}      \\
		ShanghaiTech \cite{RevisitSparse2017luoa}   				& 2017 & 330 / 275k       & 107 / 43k      & 13  & 856$\times$480   & Bikers, loitering, fighting, vehicles, \textit{etc.}          \\
		IITB Corridor \cite{MultitimescaleTrajectory2020rodriguesa} & 2020 & 208 / 302k       & 150 / 182k     & 1   & 1920$\times$1080 & Protest, playing with ball, unattended baggage, \textit{etc.} \\ \bottomrule

	\end{tabular}
\end{table*}

We evaluate our method on four commonly used datasets, \textit{i.e.} ShanghaiTech \cite{RevisitSparse2017luoa}, CUHK Avenue \cite{AbnormalEvent2013lua}, IITB Corridor \cite{MultitimescaleTrajectory2020rodriguesa} and UCSD Ped2 \cite{AnomalyDetection2010mahadevan}, which are shown in \tablename~\ref{tab2}.
ShanghaiTech is an extremely challenging dataset, since there are 13 separate scenes and the anomalous events vary widely.
We need to train only one model to detect the abnormal events in all scenes.
Although Avenue is a single-scene dataset, the complex pedestrian movements in the background making it difficult to detect anomalies.
Corridor is a newly proposed challenging dataset which has a high resolution. It contains group-level anomalies, \textit{e.g.} protest, which are absent in other datasets.
Ped2 is a small-scale dataset and all of the abnormal events are related to objects. All the frames in Ped2 are grayscale and with low resolution.
We mainly use ShanghaiTech and Avenue for ablation studies, and compare with other methods on the four datasets.

\subsection{Evaluation Metric}
We adopt the most widely used area under curve (AUC) to evaluate the performance of anomaly detection.
AUC is computed by the area under the receiver operating characteristic (ROC) curve, which is drawn by false positive rates and true positive rates with changing the threshold of anomaly scores.
In order to compare with existing methods fairly, we adopt both micro-AUC and macro-AUC following \cite{BackgroundAgnosticFramework2021georgescua}.
Micro-AUC is obtained by concatenating all frames in a dataset as a video and calculating the AUC.
Macro-AUC is the average AUC of all videos.
In ablation studies, we only report micro-AUC, which is adopted in most of the previous works.

\subsection{Implementation Details}
The default settings in our experiments are introduced as follows.
In the context recovery stream, the temporal sampling rate is set to 2, and the frames are resized to 256$\times$256 pixels before fed into STU-Net.
To mitigate the serious distortion cause by resizing, we manually crop three fixed square regions along the corridor\footnote{The left-top corners and widths are (x, y, w) = \{(320, 184, 896), (576, 96, 412), (672, 0, 256)\}}.
The weight of L1 loss $\lambda_{L1}$ in Eq. (\ref{eq2}) is set to 1.
The predefined sizes of sliding windows (\textit{i.e.} $K$s) are $\{2^{n}\}_{n=4}^{8}$, and the final sizes used in our experiments on ShanghaiTech, Avenue, Corridor and Ped2 are 128, 64, 256 and 32, respectively.
In the knowledge retrieval stream, each input snippet consists of 8 frames.
The temporal sampling rates are 8 for ShanghaiTech and Corridor, and 4 for Avenue and Ped2.
We use $B=8$ hash layers in the hash encoder, and the length of each hash code $R$ is $32$.
The weights $\lambda_m$, $\lambda_{cr}$ and $\lambda_{kr}$ are set to 0.64, 1 and 1, respectively.
The event encoder I3D is pre-trained on Kinetics-400 dataset \cite{QuoVadis2017carreira, fan2020pyslowfast} and freezed during training.
Following previous works \cite{FutureFrame2021luo, ObjectCentricAutoEncoders2019ionescua, AnomalyDetection2021georgescua, LearningNormal2021lva, BackgroundAgnosticFramework2021georgescua}, we normalize the anomaly scores and apply a Gaussian filter to smooth them.
All the experiments are performed with two Nvidia Tesla V100 GPUs using PyTorch \cite{AutomaticDifferentiation2017paszke}.
More details can be seen in our code: \url{https://github.com/zugexiaodui/TwoStreamUVAD}.

\subsection{Ablation Study}
\begin{table*}[!t]
	\centering
	\caption{Effect of Different Components}
	\label{tab3}
	\begin{tabular}{@{}ccccccccccc@{}}
		\toprule
		& \multicolumn{4}{c}{Context   recovery stream}       & \multicolumn{4}{c}{Knowledge   retrieval stream}  & \multicolumn{2}{c}{Micro-AUC   (\%)} \\ \cmidrule(l){2-11} 
		index & U-Net      & STU-Net    & Pre-training & MLE  & iL\textsuperscript{2}SH  & Training   & w/ neg.    & w/ $\lambda_m$ & ShanghaiTech  & Avenue  \\ \midrule
		1     & \checkmark &            & \checkmark   &            &            &            &            &            & 72.5                 & 82.0          \\
		2     &            & \checkmark &              &            &            &            &            &            & 74.4                 & 84.8          \\
		3     &            & \checkmark & \checkmark   &            &            &            &            &            & 75.3                 & 85.1          \\
		4     & \checkmark &            & \checkmark   & \checkmark &            &            &            &            & 75.8                 & 85.1          \\
		5     &            & \checkmark & \checkmark   & \checkmark &            &            &            &            & 79.7                 & 87.2          \\ \midrule
		6     &            &            &              &            & \checkmark &            &            &            & 71.8                 & 83.2          \\
		7     &            &            &              &            & \checkmark & \checkmark & \checkmark &            & 78.9                 & 86.0          \\
		8     &            &            &              &            & \checkmark & \checkmark &            &            & 79.6                 & 86.7          \\
		9     &            &            &              &            & \checkmark & \checkmark &            & \checkmark & 81.0                 & 88.1          \\ \midrule
		1+6   & \checkmark &            & \checkmark   &            & \checkmark &            &            &            & 75.4                 & 86.7          \\
		1+9   & \checkmark &            & \checkmark   &            & \checkmark & \checkmark &            & \checkmark & 80.2                 & 88.2          \\
		5+6   &            & \checkmark & \checkmark   & \checkmark & \checkmark &            &            &            & 81.5                 & 88.9          \\
		5+9   &            & \checkmark & \checkmark   & \checkmark & \checkmark & \checkmark &            & \checkmark & 83.7                 & 90.8          \\ \bottomrule
	\end{tabular}
\end{table*}

\begin{table}[!t]
	\centering
	\caption{Fusion of Different Methods on ShanghaiTech (Micro-AUC \%)}
	\label{tab4}
	\setlength\tabcolsep{5pt}
	\begin{tabular}{c|ccc|ccc}
		\hline
		& \multicolumn{3}{c|}{Context   recovery methods} & \multicolumn{3}{c}{Knowledge   retrieval methods} \\ \cline{2-7} 
		Method   & STU-Net        & MPN           & AMMC          & iL\textsuperscript{2}SH           & CAC            & Exemplar       \\ \hline
		STU-Net  & \textcolor[rgb]{0.5,0.5,0.5}{79.7}  & 77.5          & 78.4          & \textbf{83.7}            & 82.1           & 80.9           \\
		MPN      & 77.5           & \textcolor[rgb]{0.5,0.5,0.5}{73.1} & 74.2          & 79.9            & 78.7           & 77.4           \\
		AMMC     & 78.4           & 74.2          & \textcolor[rgb]{0.5,0.5,0.5}{73.7} & 80.4            & 79.3           & 77.9           \\ \hline
		iL\textsuperscript{2}SH    & \textbf{83.7}           & 79.9          & 80.4          & \textcolor[rgb]{0.5,0.5,0.5}{81.0}   & 80.9           & 79.1           \\
		CAC      & 82.1           & 78.7          & 79.3          & 80.9            & \textcolor[rgb]{0.5,0.5,0.5}{75.8}  & 77.9           \\
		Exemplar & 80.9           & 77.4          & 77.9          & 79.1            & 77.9           & \textcolor[rgb]{0.5,0.5,0.5}{74.2}  \\ \hline
	\end{tabular}
\end{table}

We conduct ablation experiments on the proposed two-stream framework to study the effect of different components.
The results on ShanghaiTech and Avenue datasets are shown in \tablename~\ref{tab3}.
In the context recovery stream, we adopt a 2D U-Net as the baseline, in which the encoder is replaced with a ResNet-50 \cite{DeepResidual2016he} network and other layers are the same as those in STU-Net.
The encoder is pre-trained on ImageNet \cite{ImageNetLarge2015russakovsky} and freezed during training.
Although the proposed STU-Net has better performance than the 2D U-Net with pre-training, it can use the parameters trained on Kinetics-400 dataset to further improve the performance.
In the knowledge retrieval stream, iL\textsuperscript{2}SH without training is adopted as the baseline.
"w/ neg." means that iL\textsuperscript{2}SH is trained with negative pairs, where the videos of negative instances differs from those of positive instances.
We take the negative cosine distance (\textit{i.e.} negative value of Eq. (\ref{eq4})) between negative pairs as the loss function and set its weight to 0.5 to be added to the loss of positive pairs $L_c$.
"w/ $L_m$" denotes iL\textsuperscript{2}SH is trained with our proposed mutual difference loss.

For convenience, we add index in \tablename~\ref{tab3} to refer to different ablation study settings.
Comparing experiment (\textit{abbr.} exp.) 2 with exp.\,1, it can be seen that our proposed STU-Net which can utilize the motion of input snippet outperforms 2D U-Net by about 2\% and 3\% on ShanghaiTech and Avenue, even if the encoder of STU-Net is not pre-trained.
Exp.\,3 indicates the I3D encoder in our STU-Net can also benefit from the learned representation in action recognition, which is an advantage compared with other models.
Equipped with MLE, the performances of U-Net and STU-Net improve by 2\%$\sim$4\% in exp.\,4 and exp.\,5, demonstrating the effectiveness of the proposed MLE.
In exp.\,8, our iL\textsuperscript{2}SH trained without negative pairs boosts the performance of the basic iL\textsuperscript{2}SH in exp.\,6 by 7.8\% and 3.5\% on ShanghaiTech and Avenue, respectively.
It also surpasses the iL\textsuperscript{2}SH trained with negative samples in exp.\,7 by 0.7\%.
With the proposed mutual difference loss in exp.\,9, the results of iL\textsuperscript{2}SH on ShanghaiTech and Avenue increase by 1.4\% compared with exp.\,8 which does not have a constraint on the difference of hash layers.
Moreover, we re-implement LLSH \cite{LearnableLocalitySensitive2021lu} on ShanghaiTech and Avenue datasets. The micro-AUCs of LLSH on the two datasets are 78.7\% and 86.3\%, which are inferior to our iL\textsuperscript{2}SH by about 2\%.
From exp.\,"1+6", we can see that the fusion of two baseline models can bring a significant improvement of 3\%$\sim$5\% compared with a single model.
Through fusing the two streams proposed in this work, our two-stream framework achieves the best results on both datasets in exp.\,"5+9".

\subsection{Complementarity Study}

\begin{table}[!t]
	\centering
	\caption{Fusion of Different Methods on Avenue (Micro-AUC \%)}
	\label{tab5}
	\setlength\tabcolsep{4.6pt}
	\begin{tabular}{c|ccc|ccc}
		\hline
		& \multicolumn{3}{c|}{Context recovery methods} & \multicolumn{3}{c}{Knowledge retrieval methods} \\ \cline{2-7} 
		Method   & STU-Net        & MNAD          & AMMC          & iL\textsuperscript{2}SH           & CAC            & Exemplar       \\ \hline
		STU-Net  & \textcolor[rgb]{0.5,0.5,0.5}{87.2}  & 88.1          & 86.8          & \textbf{90.8}            & 88.3           & 88.3           \\
		MNAD     & 88.1           & \textcolor[rgb]{0.5,0.5,0.5}{87.5} & 88.4          & 90.6            & 89.7           & 89.2           \\
		AMMC     & 86.8           & 88.4          & \textcolor[rgb]{0.5,0.5,0.5}{86.6} & 90.3            & 89.6           & 88.9           \\ \hline
		iL\textsuperscript{2}SH    & \textbf{90.8}           & 90.6          & 90.3          & \textcolor[rgb]{0.5,0.5,0.5}{88.1}   & 87.0           & 87.3           \\
		CAC      & 88.3           & 89.7          & 89.6          & 87.0            & \textcolor[rgb]{0.5,0.5,0.5}{83.0}  & 84.3           \\
		Exemplar & 88.3           & 89.2          & 88.9          & 87.3            & 84.3           & \textcolor[rgb]{0.5,0.5,0.5}{84.1}  \\ \hline
	\end{tabular}
\end{table}

To verify the complementarity of context recovery and knowledge retrieval, we re-implement several recent methods, each of which can be generalized as a kind of context recovery or knowledge retrieval method, to make a trough fusion study.
The results on ShanghaiTech and Avenue datasets are shown in ~\tablename~\ref{tab4} and ~\tablename~\ref{tab5}.
The micro-AUCs of basic methods which are not fused with others are shown on the diagonal and in gray text.
An off-diagonal value denotes the AUC of fusing the methods corresponding to its column and its row.
The highest results are shown in bold.
Since the weights for fusing two streams are both set to 1, the result matrix in each table is symmetric.

In addition to our STU-Net, the context recovery methods include MPN \cite{LearningNormal2021lva}, MNAD \cite{LearningMemoryGuided2020parka} and AMMC \cite{AppearanceMotionMemory2021caia}, which are state-of-the-art memory-augmented context recovery models.
We follow their official codes for re-implementation.
The re-implemented knowledge retrieval methods are CAC \cite{ClusterAttention2020wanga} and Exemplar Selection \cite{StreetScene2020ramachandraa} (\textit{abbr.} Exemplar), which have been introduced in Related Work.
For these two methods, we use the same pre-trained I3D encoder as in iL\textsuperscript{2}SH for feature extraction.
In CAC, we freeze the pre-trained feature extractor and only train the cluster attention module instead of training the whole network.
We report the result under the setting of 16 clusters since it achieves the best performance.
As to Exemplar, we adopt MSE to measure the distance between two samples, and the distance thresholds for constructing exemplar sets are set to 150 and 60 for ShanghaiTech and Avenue datasets, respectively.
We take the average distance between the testing sample and its 8 / 64 nearest exemplars for ShanghaiTech / Avenue as the anomaly score, which achieves the best performance compared with other settings.

From \tablename~\ref{tab4} and \tablename~\ref{tab5}, we can see that the fusion of two context recovery methods or two knowledge retrieval methods cannot bring an obvious improvement.
For example, in \tablename~\ref{tab4}, by fusing MPN (73.1\%) and AMMC (73.7\%), the AUC (74.2\%) is only increased by 0.5\% compared with the higher performance between MPN and AMMC (\textit{i.e.} 73.7\%).
However, fusing two different types of methods can generally brings a significant improvement, even though the context recovery methods are equipped with memory modules.
For example, fusing CAC and AMMC has an improvement of 3.5\% (75.8\% + 73.7\% $\rightarrow$ 79.3\%) on ShanghaiTech.
In some cases, the fusion of a context recovery method and a knowledge retrieval method may have a slight decline, which is reasonable since the results of the two methods have a huge gap.

\begin{figure}[!t]
	\centering
	\includegraphics[]{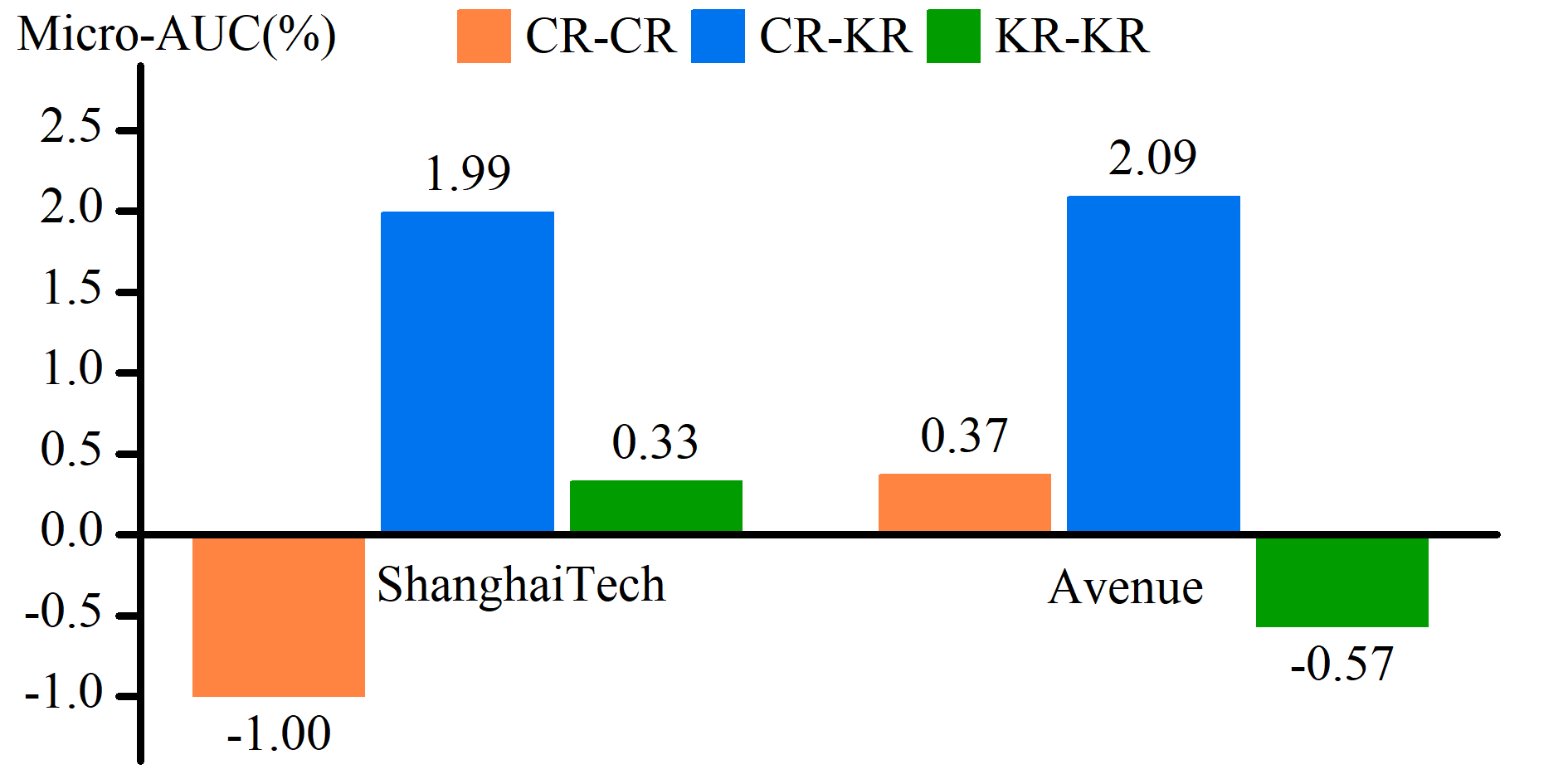}
	\caption{Performance variations of fusing different types of methods (CR: context recovery; KR: knowledge retrieval).}
	\label{fig_fusion_type}
\end{figure}

\begin{table}[!t]
	\centering
	\caption{Fusion of Two Streams Which Have the Same Temporal Sampling Rate (=4)}
	\label{tab6}
	\begin{tabular}{@{}cccc@{}}
		\toprule
		& \multicolumn{3}{c}{Micro-AUC (\%)} \\ \cmidrule(l){2-4} 
		Method    					& ShanghaiTech    & Avenue    & Ped2   \\ \midrule
		STU-Net    					& 77.3            & 84.6      & 90.0   \\
		iL\textsuperscript{2}SH     & 78.7      	  & 88.1      & 91.3   \\
		Two-Stream 					& 81.1            & 88.8      & 93.4   \\
		Improvement 				& 2.4             & 0.7       & 2.1    \\ \bottomrule
	\end{tabular}
\end{table}

To analyze the effect of fusing different types of methods clearly, we calculate the average improvement of each fusion type based on the results in \tablename~\ref{tab4} and \tablename~\ref{tab5}.
The performance variations are shown in \figurename~\ref{fig_fusion_type}, where the fusion types include two context recovery methods (CR-CR), two knowledge retrieval methods (KR-KR) and a context recovery method with a knowledge retrieval method (CR-KR).
It can be seen that the fusion of context recovery methods and knowledge retrieval methods can bring the highest improvement on both datasets, which significantly exceeds the fusion of two identical types of methods by more than 1.7\%, demonstrating the complementarity of context recovery and knowledge retrieval.
Particularly, when STU-Net and iL\textsuperscript{2}SH are fused, the results are the highest on both ShanghaiTech and Avenue datasets as displayed in \tablename~\ref{tab4} and \tablename~\ref{tab5}.

Furthermore, we conduct an experiment where the STU-Net and iL\textsuperscript{2}SH have the same temporal sampling rate, to verify that the improvement of fusing a context recovery stream and a knowledge retrieval stream is not caused by different temporal scales.
As shown in \tablename~\ref{tab6}, the sampling rates of STU-Net and iL\textsuperscript{2}SH are both set to 4.
Although this setting results in lower performance compared with the default setting, the improvements on ShanghaiTech, Avenue and Ped2 datasets are still significant.
This experiment proves that fusing the context recovery method and knowledge retrieval method which have the same temporal scale can bring improvement for the fusion.
On the contrary, even if two methods of the same type have different temporal scales, the fusion of them cannot boost the performance.
For example, the context recovery methods STU-Net and AMMC in \tablename~\ref{tab5} have different temporal sampling rates (2 and 1), the result of fusion is lower than STU-Net by 0.4\%.

To sum up, context recovery methods and knowledge retrieval methods can complement each other and bring significant improvement while the same type of methods cannot.
Among the fusions of different methods, our proposed two-stream model consisting of STU-Net and iL\textsuperscript{2}SH achieves the best performance, which demonstrates the superiority of the proposed method.

\subsection{Visualization and Analysis}
We visualize several testing samples and anomaly scores to analyze the effects of MLE and fusion of STU-Net and iL\textsuperscript{2}SH in this section.

\begin{figure}[!t]
	\centering
	\includegraphics[]{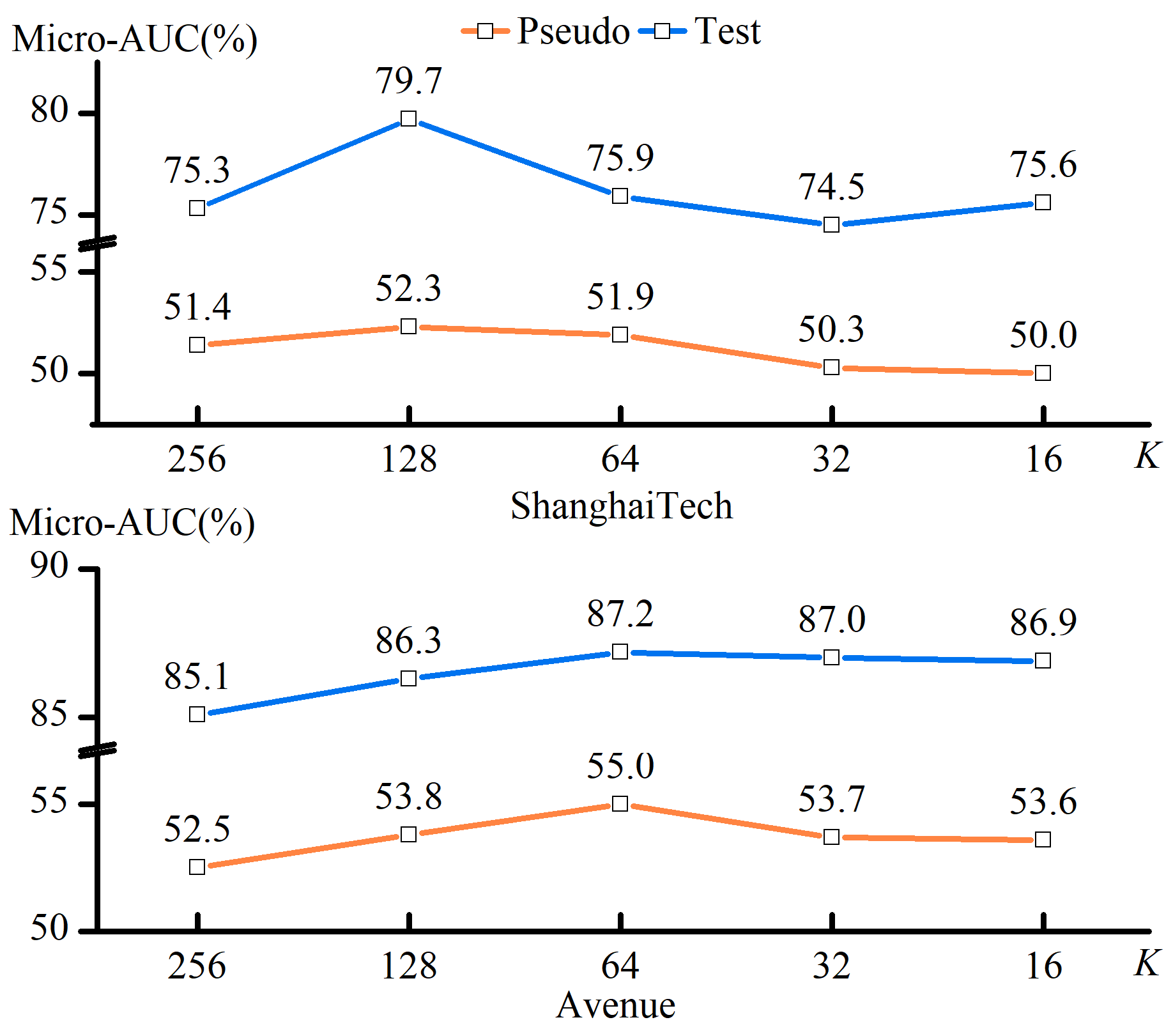}
	\caption{AUCs on pseudo anomalous data and real testing data. $K$ denotes the size of sliding window in MLE.}
	\label{fig_mle_curve}
\end{figure}

\begin{figure}[!t]
	\centering
	\includegraphics[]{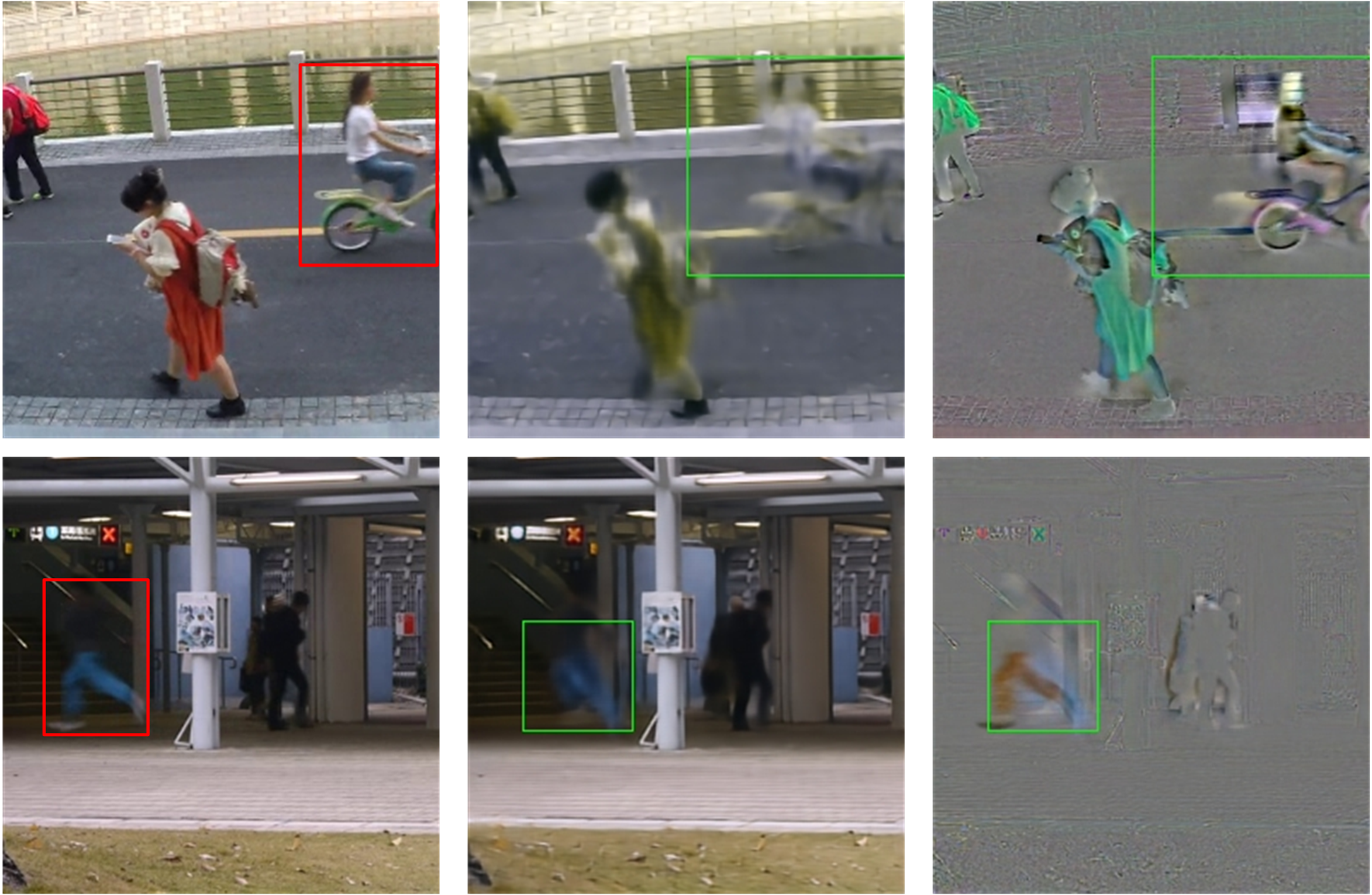}
	\caption{Visualization of frames from ShanghaiTech (the first row) and Avenue (the second row) datasets. The columns are ground truth frames, predicted frames and error frames. Red boxes represent the anomalous regions, and green boxes denote the patches which have the maximum error.}
	\label{fig_mle_sample}
\end{figure}

\begin{figure}[!t]
	\centering
	\includegraphics[]{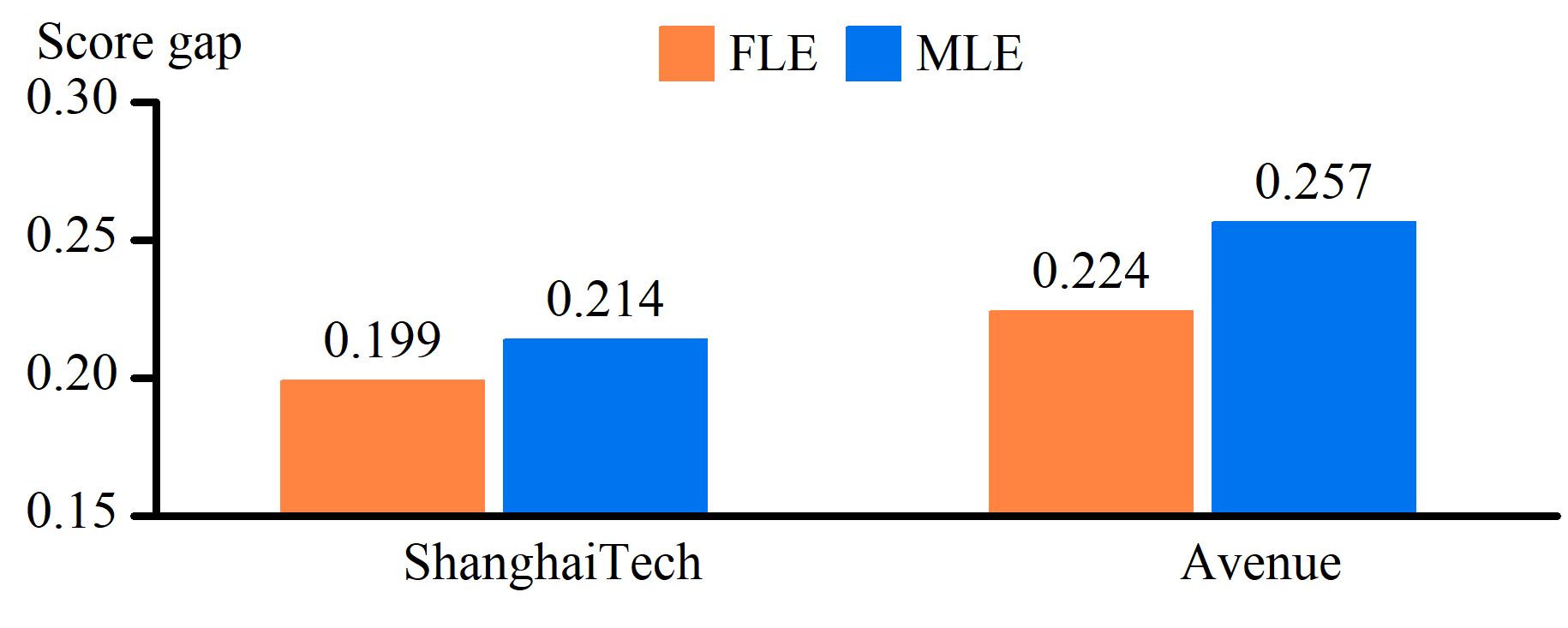}
	\caption{Score gaps of frame-level error (FLE) and maximum local error (MLE).}
	\label{fig_mle_gap}
\end{figure}

\subsubsection{Maximum Local Error}

\figurename~\ref{fig_mle_curve} displays the AUCs of different sizes of the sliding window (\textit{i.e.} $K$s) in MLE on testing data and pseudo anomalous data simulated by training videos.
On both ShanghaiTech and Avenue datasets, it can be seen the AUC curves on the two types of data almost have the same trend.
On ShanghaiTech dataset, the highest AUC on pseudo anomalous data is achieved when $K=128$, according to which we set $K$ to 128 on testing data and achieves the best result.
The same is true on Avenue dataset when $K=64$, which demonstrates the effectiveness of simulating anomalies.

Two examples of frames are shown in \figurename~\ref{fig_mle_sample}, from which we can see MLE finds the anomalous region accurately without using any object detection algorithms.
To quantitatively study how MLE affects anomaly scores, we calculate the score gaps of frame-level error (FLE) and MLE, as shown in \figurename~\ref{fig_mle_gap}.
The score gap is the difference value between the average scores of anomalous frames and normal frames.
It reflects the ability to discriminate normality and anomaly, and is expected to has a higher value.
\figurename~\ref{fig_mle_gap} shows that MLE can increase the score gap, thus improving the ability to detect anomalies in videos.

\subsubsection{Fusion of Two Streams}
\begin{figure}[!t]
	\centering
	\includegraphics[]{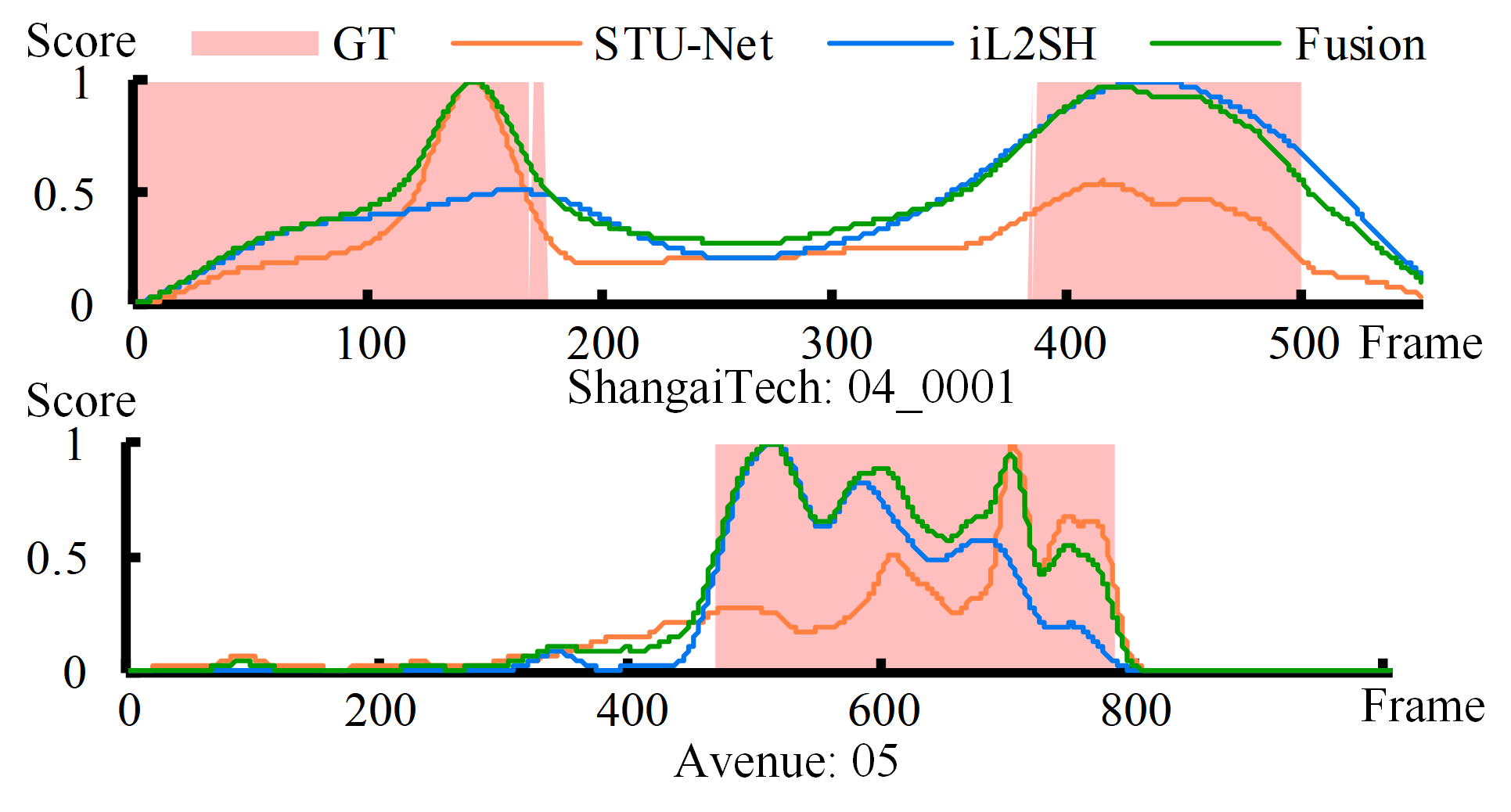}
	\caption{Anomaly score curves of STU-Net, iL\textsuperscript{2}SH and fusion of the two streams on two videos. GT: ground truth.}
	\label{fig_fusion_score_curve}
\end{figure}

\begin{figure}[!t]
	\centering
	\includegraphics[]{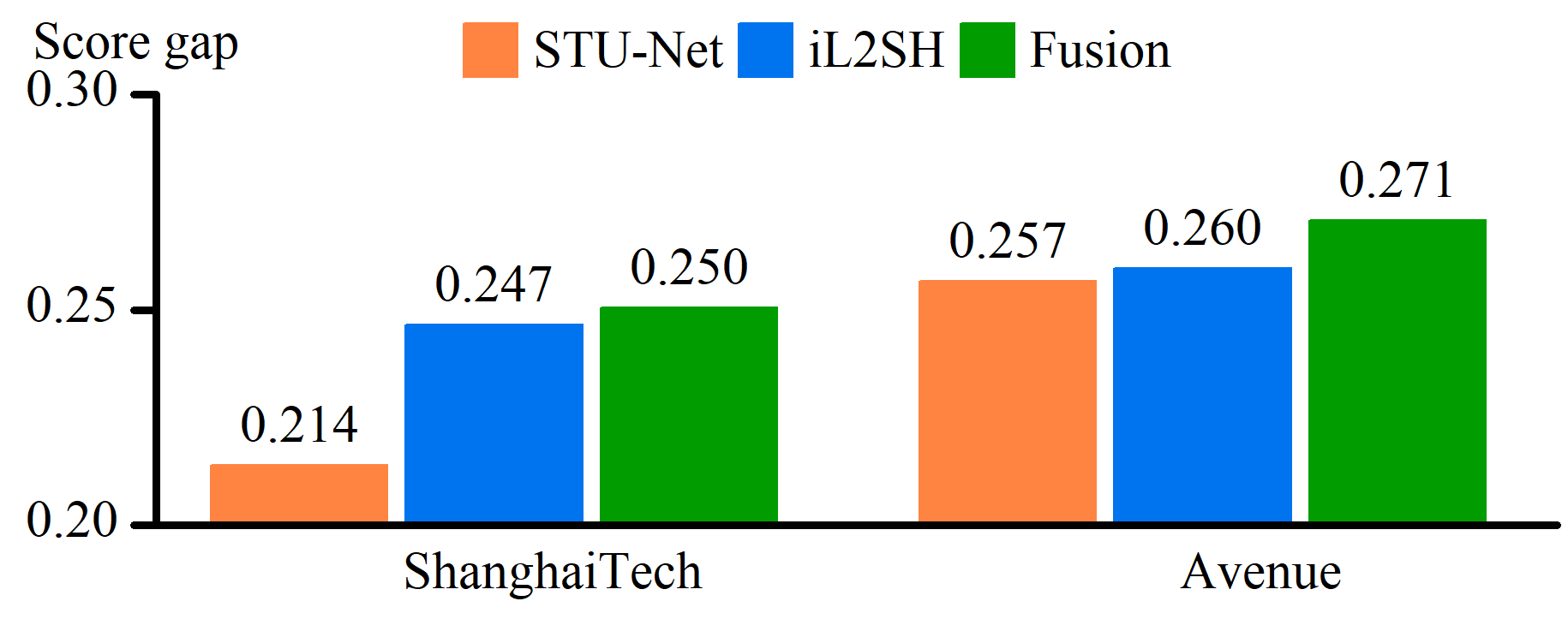}
	\caption{Score gaps of STU-Net, iL\textsuperscript{2}SH and fusion of the two streams.}
	\label{fig_fusion_gap}
\end{figure}

We illustrate the anomaly score curves of each stream and fusion of the two streams in \figurename~\ref{fig_fusion_score_curve} to explain how the two streams complement each other.
For example, in video "04\_0001" of ShanghaiTech, iL\textsuperscript{2}SH and STU-Net output relative low anomaly scores in the first and second abnormal periods, respectively.
However, the fusion of the two streams can generate high anomaly scores in both periods, hence improving the AUC by 2.7\% on this video.
The score gaps are displayed in \figurename~\ref{fig_fusion_gap}, which shows that the fusion of the two streams increases the score gap.
Therefore, our two-stream framework can achieve better performance than a single stream.

\subsection{Comparison with Existing Methods}

\newcommand\ts{\textsuperscript}
\newcommand\tdar{\textdagger}
\newcommand\tddr{\textdaggerdbl}
\newcommand\cmk{\checkmark}
\newcommand\tgy{\textcolor[rgb]{0.5,0.5,0.5}}

\begin{table*}[!t]
	\centering
	\caption{Comparison of Different Methods on Four Datasets. Text in Bold: the Best Result; \ts{\tdar}: Results Implemented by Others; \ts{\tddr}: Using Manually Cropped Regions; w/ obj.: Using Object Detection; Underline: the Best Result wo/ Using Object Detection}
	\label{tab:sota}
	\begin{tabular}{@{}lccccccccc@{}}
		\toprule
		&      	& \multicolumn{2}{c}{ShanghaiTech} 		& \multicolumn{2}{c}{Avenue}& \multicolumn{2}{c}{Corridor} 	& \multicolumn{2}{c}{Ped2} \\\cmidrule(l){3-10} 
		Method     											&w/ obj.& Micro           	& Macro          	& Micro		& Macro       	& Micro         & Macro        	& Micro     & Macro      \\ \midrule
		FFP\cite{FutureFrame2018liua}  						&    	& 72.8            	& -              	& 84.9		& -				&~64.7\ts{\tdar}& - 		& 95.4 		& -   \\
		MemAE\cite{MemorizingNormality2019gonga}			&    	& 71.2            	& -              	& 83.3      & -				& -				& -            	& 94.1		& -          \\
		OADA\cite{ObjectCentricAutoEncoders2019ionescua}	& \cmk  & -		        	& 84.9       		& 			& 90.4			& -				& -      		& -			& 97.8 \\
		MPED-RNN\cite{LearningRegularity2019moraisa}		&    	& 73.4            	& -              	& -			& -				&~64.3\ts{\tdar}& - 			& -			& -          \\
		AMC\cite{AnomalyDetection2019nguyena}				&    	& -	            	& -              	& 86.9		& -				& -				& -            	& 96.2		& -          \\
		BMAN\cite{BMANBidirectional2020lee}					&    	& 76.2            	& -              	& 90.0		& -				& -				& -            	& 96.6		& -          \\
		r-GAN\cite{FewShotSceneAdaptive2020lua}				&    	& 77.9            	& -              	& 85.8		& -				& -				& -            	& 96.2		& -          \\
		GEPC\cite{GraphEmbedded2020markovitza}				&    	& 76.1            	& -              	& -			& -				& -				& -            	& -			& -          \\
		MNAD\cite{LearningMemoryGuided2020parka}			&    	& 70.5            	& -              	& 88.5		& -				& -				& -            	& 97.0		& -          \\
		MTP\cite{MultitimescaleTrajectory2020rodriguesa}	&    	& 76.0            	& -              	& 82.9		& -				& 67.1			& -            	& -			& -          \\
		Ada-Net\cite{LearningNormal2020song} 				&    	& 70.0            	& -              	& 89.2		& -				& -				& -            	& 90.7		& -          \\
		CAC\cite{ClusterAttention2020wanga}					&    	& 79.3            	& -              	& 87.0		& -				& -				& -            	& -			& -          \\
		DeepOC\cite{DeepOneClass2020wu} 					&    	& -            		& -              	& 86.6		& -				& -				& -            	& 96.9		& -          \\
		VEC-AM\cite{ClozeTest2020yu} 						& \cmk 	& 74.8       		& -          		& 89.6		& -				& -				& -      		& 97.3		& -    		\\
		Multispace\cite{NormalityLearning2021zhang}			&    	& 73.6            	& -              	& 86.8		& -				& -				& -            	& 95.4		& -          \\
		AMMC-Net\cite{AppearanceMotionMemory2021caia}		&    	& 73.7           	& -              	& 86.6		& -				& -				& -            	& 96.6		& -          \\
		MESDnet\cite{MultiEncoderEffective2021fang}			&    	& 73.2            	& -              	& 86.3		& -				& -				& -            	& 95.6		& -          \\
		BAF\cite{BackgroundAgnosticFramework2021georgescua} & \cmk 	& 82.7				& 89.3				& \textbf{92.3}	& 90.4		& -				& -      		& 98.7		& 99.7 \\
		SSMTL\cite{AnomalyDetection2021georgescua}			& \cmk 	& -          		& 90.2       		& -			& 92.8			& -				& -      		& -			& \textbf{99.8} \\
		HF\ts{2}-VAD\cite{HybridVideo2021liua}				& \cmk 	& 76.2       		& -          		& 91.1		& -				& -				& -      		& \textbf{99.3}& -      \\
		F\ts{2}PN\cite{FutureFrame2021luo}					&    	& 73.0            	&              		& 85.7		& -				& -				&             	& 96.2		& -          \\
		LLSH\cite{LearnableLocalitySensitive2021lu}			&    	& 77.6            	& 85.9              & 87.4		& 88.6			&~73.5\ts{\tddr}&~74.2\ts{\tddr}& -			& -          \\
		sRNN-AE\cite{VideoAnomaly2021luo}					&    	& 69.6            	& -              	& 83.5		& -				& -				& -            	& 92.2		& -          \\
		MPN\cite{LearningNormal2021lva}						&    	& 73.8            	& -              	& 89.5		& -				& -				& -            	& 96.9		& -          \\
		SmithNet\cite{SmithNetStrictness2021nguyen}			&    	& 73.8            	& -              	& 89.4		& -				& -				& -            	& \underline{98.4}& -          \\
		ROADMAP\cite{RobustUnsupervised2021wanga}			&    	& 76.6            	& -              	& 88.3		& -				& -				& -            	& 96.3		& -          \\
		HSTGCNN\cite{HierarchicalSpatioTemporal2021zeng}	& \cmk 	& 81.8       		& -           		& 87.5		& -				& 70.5			& -		      	& 97.7		& -		    \\
		SIGnet\cite{AnomalyDetection2022fang}				&    	& -	            	& -              	& 86.8		& -				& -				& -            	& 96.2		& -          \\
		SSAGAN\cite{SelfSupervisedAttentive2022huang}		&    	& 74.3            	& -              	& 88.8		& -				& -				& -            	& 97.6		& -          \\
		VABD\cite{VariationalAbnormal2022li}				&    	& 78.2            	& -              	& 86.6		& -				& 72.2			& -            	& 97.1		& -          \\ \midrule
		STU-Net (ours)										&       & 79.7				& 87.6          	& 87.2      & 88.2          &~74.9\ts{\tddr}&~77.2\ts{\tddr}& 95.9     	& 97.4      \\
		iL\textsuperscript{2}SH (ours)						&       & 81.0          	& 87.2          	& 88.1      & 90.6         	&~68.8\ts{\tddr}&~65.6\ts{\tddr}& 91.3     	& 99.2      \\
		Two-stream (ours)									&       & \textbf{\underline{83.7}} 	& \textbf{\underline{90.8}} 	& \underline{90.8}      & \textbf{\underline{93.0}} & ~\textbf{\underline{78.3}}\ts{\tddr} & ~\textbf{\underline{77.9}}\ts{\tddr}	& 97.1    	& \underline{99.3}      \\ \bottomrule
	\end{tabular}
\end{table*}

\begin{table}[!t]
	\centering
	\caption{Results on Corridor without manually cropping}
	\label{tab:cor480}
	\begin{tabular}{@{}lcc@{}}
		\toprule
		& \multicolumn{2}{c}{Corridor} \\ \cmidrule(l){2-3} 
		Method     					& Micro         & Macro        \\ \midrule
		STU-Net    					& 69.5         	& 63.6        \\
		iL\textsuperscript{2}SH		& 70.7        	& 59.7        \\
		Two-stream 					& 73.1        	& 64.0        \\ \bottomrule
	\end{tabular}
\end{table}

The comparison of different methods on ShanghaiTech \cite{RevisitSparse2017luoa}, Avenue \cite{AbnormalEvent2013lua}, Corridor \cite{MultitimescaleTrajectory2020rodriguesa} and Ped2 \cite{AnomalyDetection2010mahadevan} datasets is displayed in \tablename~\ref{tab:sota}.
We report both micro-AUC and macro-AUC for each method if available, and the best results are in bold text.
The results marked with \tdar\, are implemented by Rodrigues \textit{et al.} \cite{MultitimescaleTrajectory2020rodriguesa} because there are no official results.
In LLSH \cite{LearnableLocalitySensitive2021lu} and our two-stream framework, the results on Corridor are under the setting of manually cropped regions aforementioned in Implementation Details, which are marked with \tddr.
We mark the methods using object detection (w/ obj.) since anomalies in most datasets are closely related to objects in the current stage of video anomaly detection.
The best results of the methods without using object detection are underlined.

Under a fair setting that does not use object detection, our two-stream method achieves the best performance in both micro-AUC and macro-AUC metrics on three datasets, \textit{i.e.} ShanghaiTech, Avenue and Corridor.
Especially, it is worthy noting that the micro-AUC of our method on ShanghaiTech dataset is higher than other two outstanding methods without using object detection, \textit{i.e.} CAC \cite{ClusterAttention2020wanga} and VADB \cite{VariationalAbnormal2022li}, by 4.4\% and 5.5\%.
Even though compared with those methods using object detection \cite{ObjectCentricAutoEncoders2019ionescua, ClozeTest2020yu, BackgroundAgnosticFramework2021georgescua, AnomalyDetection2021georgescua, HybridVideo2021liua, HierarchicalSpatioTemporal2021zeng}, pose estimation \cite{HierarchicalSpatioTemporal2021zeng} and extra datasets \cite{BackgroundAgnosticFramework2021georgescua} (served as pseudo anomalous data), our method is still the best on the two large-scale datasets (\textit{i.e.} ShanghaiTech and Corridor) in both metrics, and the best on Avenue in macro-AUC.
We also experiment on Corridor without manually selecting regions as shown in \tablename~\ref{tab:cor480}, and our two-stream framework still achieves the best performance compared with other methods \cite{FutureFrame2018liua, LearningRegularity2019moraisa, MultitimescaleTrajectory2020rodriguesa, HierarchicalSpatioTemporal2021zeng, VariationalAbnormal2022li}.
Although the performance of our model is not the best on Ped2 dataset, it can be improved by combining with object detection and optical flow in the future since all the anomalies in Ped2 are related to objects.

\section{Conclusion}
In this paper, we propose a novel two-stream framework composed of a context recovery stream and a knowledge retrieval stream for video anomaly detection.
In the context recovery stream, a spatiotemporal U-Net (STU-Net) is proposed to utilize the motion in the current snippet to predict the future frame.
Additionally, we propose a maximum local error (MLE) mechanism which can focus on the recovery error in anomalous region and hence generate more accurate anomaly score.
In the knowledge retrieval stream, we propose an improved learnable locality-sensitive hashing (iL\textsuperscript{2}SH) to store the knowledge about normality and retrieve it to determine whether a testing event is consistent with the normal knowledge.
By fusing the context recovery stream and the knowledge retrieval stream, our two-stream framework can use both short-term motion and knowledge about normality to detect anomalies.
Extensive experiments verify the effectiveness and complementarity of the two streams, which achieves the state-of-the-art performance on ShanghaiTech, Avenue, Corridor and Ped2 datasets.


\bibliographystyle{IEEEtran}
\bibliography{IEEEabrv,mybib}

\newpage


\end{document}